# Artificial intelligences and human scientists exhibit complementary strengths in theory building

Ke Li[1], Spyros I. Zoumpoulis[1], Phanish Puranam[1], Philip Parker[1], Matthew Eshbaugh-Soha[2], Izzy Gainsburg[3], Michael Gilead[4], Igor Grossmann[5], Britt Hadar[6], Yoel Inbar[7], Almog Simchon[8], Robb Willer[3], Rui Ai[9], Ruicheng Ao[9], Gavin J. Bala[1], Matthew Bidwell[10], Shuang Cai[11], Kai Chang[9], Skyler Y. Chen[12], Cory J. Clark[13], Irmak Dai[1], Abhinandan Dalal[14], Connor Douglas[15], Alexis Du[1], Zhehang Du[10], Leyun Feng[16], Isabel Fernandez-Mateo[17], Linnea Gandhi[10], Cyrille Grumbach[18], Anmol Gupta[3], Vansh Gupta[9], Maria Hademer[1], Jay H. Hardy III[19], Chen Kai Huang[16], Jacob Xiangyu Jin[17], Ufuk Keskin[9], Na Hyun Kim[10], Mert Kobaş[15], Byounghoon Koh[1], Gabrielle Lamont-Dobbin[20], Gregory Lanzalotto[10], Sun Young Lee[21], Dingzhe Leng[12], Chenjun Li[11], Weiyuan Li[11], Zeyuan Li[9], Zhongyuan Liang[12,22], Ning Liu[23], Peihong Liu[24], Yuhan Liu[25], Jiuyao Lu[10], Wanteng Ma[10], Nicolas Martinet[7], Natnael Mulat[20], Christina A. Nguyen[26], Khai Nguyen[9], Quang Minh Nguyen[9], Naja Pape[27], Chanwoo Park[9], Stefanos Poulidis[1], Jeffrey Sanchez-Burks[28], Michael Schaerer[29], Isabelle Solal[30], Yanbo Song[31], Junghyo Sun[1], Qingyao Sun[11], Rui Sun[32], Roderick Swaab[1], Kevin Tan[10], Dequn Teng[25], Michelle A. Vaccaro[14], Robin Vigerbaeck[1], Xiaomeng Wang[10], Randol H. Yao[9], Duygu Yilmaz[15], Shun Yiu[33], Ecem Yucesoy[1], Allen Zang[34], Ruijia Zhang[23], Xilan Zhang[3], Yichi Zhang[15], Zhanhao Zhang[11], Eric Luis Uhlmann[1]

Corresponding author: ke.li@insead.edu

**Affiliations**

[1] INSEAD
[2] University of North Texas
[3] Stanford University
[4] Tel Aviv University
[5] University of Waterloo
[6] Reichman University
[7] University of Toronto
[8] Ben-Gurion University of the Negev
[9] Massachusetts Institute of Technology
[10] University of Pennsylvania
[11] Cornell University
[12] University of California, Berkeley
[13] New College of Florida
[14] Harvard University
[15] New York University
[16] Northwestern University
[17] London Business School
[18] ETH Zurich
[19] Oregon State University

[20] Columbia University
[21] University College London
[22] University of California, San Francisco
[23] Johns Hopkins University
[24] Emirates Aviation University
[25] University of Cambridge
[26] Fordham University
[27] Imperial College
[28] University of Michigan
[29] Singapore Management University
[30] ESSEC Business School
[31] The Hong Kong Polytechnic University
[32] Stockholm School of Economics
[33] Indiana University
[34] University of Chicago

*Author contributions.* Project ideation: Ke Li, Spyros I. Zoumpoulis, Phanish Puranam, Philip Parker, and Eric Uhlmann. Data collection: Ke Li and Philip Parker. Data analysis: Ke Li, Spyros I. Zoumpoulis, Phanish Puranam, and Philip Parker. Drafted manuscript: Ke Li, Spyros I. Zoumpoulis, Phanish Puranam, Philip Parker, and Eric Uhlmann. Matthew Eshbaugh-Soha, Izzy Gainsburg, Michael Gilead, Igor Grossmann, Britt Hadar, Yoel Inbar, Almog Simchon, and Robb Willer served on the expert panel that identified and operationalized the focal variables for the machine learning analysis of inequality discourses. Rui Ai, Ruicheng Ao, Gavin J. Bala, Matthew Bidwell, Shuang Cai, Kai Chang, Skyler Y. Chen, Cory J. Clark, Irmak Dai, Abhinandan Dalal, Connor Douglas, Alexis Du, Zhehang Du, Matthew Eshbaugh-Soha, Leyun Feng, Isabel Fernandez-Mateo, Izzy Gainsburg, Linnea Gandhi, Igor Grossmann, Cyrille Grumbach, Anmol Gupta, Vansh Gupta, Britt Hadar, Maria Hademer, Jay H. Hardy III, Chen Kai Huang, Yoel Inbar, Jacob Xiangyu Jin, Ufuk Keskin, Na Hyun Kim, Mert Kobaş, Byounghoon Koh, Gabrielle Lamont-Dobbin, Gregory Lanzalotto, Sun Young Lee, Dingzhe Leng, Chenjun Li, Weiyuan Li, Zeyuan Li, Zhongyuan Liang, Ning Liu, Peihong Liu, Yuhan Liu, Jiuyao Lu, Wanteng Ma, Nicolas Martinet, Natnael Mulat, Christina A. Nguyen, Khai Nguyen, Quang Minh Nguyen, Naja Pape, Chanwoo Park, Stefanos Poulidis, Jeffrey Sanchez-Burks, Michael Schaerer, Isabelle Solal, Yanbo Song, Junghyo Sun, Qingyao Sun, Rui Sun, Roderick Swaab, Kevin Tan, Dequn Teng, Michelle A. Vaccaro, Robin Vigerbaeck, Xiaomeng Wang, Randol H. Yao, Duygu Yilmaz, Shun Yiu, Ecem Yucesoy, Allen Zang, Ruijia Zhang, Xilan Zhang, Yichi Zhang, and Zhanhao Zhang formulated initial theories, predicted findings, and generated theoretical explanations for the observed results regarding race and gender discourses. All authors edited and commented on the manuscript draft.

*Acknowledgments.* The authors acknowledge the generous support of the INSEAD R&D committee and Human and Machine Intelligence Institute (HUMII).

# Abstract

We investigate the effectiveness of artificial intelligences (AI)—specifically large language models (LLMs)—relative to human scientists at high-level cognitive tasks in social science such as theory formulation, predictions of novel empirical results, and theory revision in response to new evidence. The research domain was academic discourse regarding gender and race inequality. Our findings, comparing 25 LLMs with 13 senior researchers and 60 doctoral scholars, reveal that the AIs outperformed most humans individually on most of the present tasks, while human theories were more diverse and exhibited greater gains in predictive accuracy from aggregation. AI-generated theories were more extensively elaborated, involving additional theoretical paths and latent variables, and were rated as higher quality than human theories by independent raters blinded to source. However, this theoretical complexity was in part ornamental, in that it was not associated with more accurate predictions about empirical patterns in data; in contrast, human scientists achieved greater predictive efficiency with simpler theories. The AIs were significantly more likely than human scientists to revise their theories to incorporate new evidence; human scientists updated their beliefs in a selective way that is sensitive to prior prediction errors. We speculate that the superior processing capacity of artificial intelligences makes them especially well-suited to tasks requiring grappling with complexity, but that the greater diversity of human ideas is essential to wise crowds and collective creativity. (Word count: 225)

**Significance statement:** We find that artificial intelligence agents perform as well, and in some cases better than, individual human scientists at a specific set of advanced theory building tasks in a social science project. AI-generated theories were structurally more complex and were perceived as higher quality. Contrarily, human scientists exhibited greater intellectual diversity, formed wiser crowds when making certain scientific predictions, and achieved higher predictive efficiency when controlling for theory complexity. These complementary comparative advantages point to promising avenues for human–AI collaboration in scientific settings.

# Introduction

Recent years have seen exponential advances in the capabilities and proliferation of artificial intelligences (AIs), including in academic scholarship, with generative large language models (LLMs) increasingly deployed as writing assistants[1], coders[2,3], transcribers[4], and data collectors[5], and simulated experimental participants[6], among other functions. Although artificial intelligences are typically used in subordinate roles in human-led teams, recent evidence from biology[7,8], mathematics[9,10], cognitive science[11,12] and computer science[13] indicate they already have utility for high-level challenges previously believed to be the purview of humans, such as theory building, hypothesis formulation, and interpretation of empirical results. But how do the ever-expanding capabilities of AIs compare to those of human scientists, both individually and collectively? And what are the comparative advantages of generative LLMs and human scholars for future human–AI scientific collaborations? In the current study, we compare the performance of human theorists and LLMs in anticipating and explaining correlational patterns in data from an important social science domain—the study of inequality discourses.

In a number of the hard sciences, artificial intelligences have shown their worth by solving previously unsolved mathematics problems[9,10], developing novel and more efficient algorithms in computer science[14,15], and discovering new materials[16,17], chemical and protein structures[7,8,18], and biochemical interactions[19]. Several recent initiatives in cognitive science[11,12,20] have "closed the loop" of scientific discovery to almost entirely exclude human researchers. AUTOmated COGnitive Scientist (AUTOCOG)[12] and Auto-Psych[11] are able to design and carry out entire experimental psychology projects largely autonomously, iteratively collecting new data with humans as subjects using online panels. Thus far, AUTOCOG and Auto-Psych have re-discovered and improved on ground-truth theoretical accounts of established phenomena such as what makes coin tosses seem random[21] and heuristics in multi-cue decision making[22]. However, in these closed-loop demonstrations, the counterfactual scenario remains unexplored: "what if the project had been led by humans?".

There are numerous reasons for skepticism regarding the ability of artificial intelligences to match or exceed the performance of human social scientists. On some accounts, LLMs are potential fonts of misinformation[23] because their underlying goal is user satisfaction and re-use, not truth[24]—potentially leading them to produce impressive-yet-false theories. Thus, it is important to tether AI ideas to empirical data to rigorously and objectively assess their quality. AIs may strain to formulate theories regarding opaque, unobservable processes in what Jagadish et al.[12] refer to as the "open conceptual space" of human behavior: psychological and social dynamics that are not bound by rigid rules or mathematical formulas. Social domains add further layers of complex interacting variables above and beyond the cognitive science-oriented topics examined in some of the major investigations on AI-driven theory discovery thus far[11,12]. The stochastic parrot argument[25] holds that LLMs merely recombine existing linguistic patterns to mimic reasoning and understanding, and are therefore inherently limited in their processes and outputs. LLM-generated outputs are stochastic, such that the boundaries of their performance on various idea-generation tasks are difficult to anticipate, again necessitating empirical tests with specific task cases.

Given the complexity of human behavior and interaction, can artificial intelligences compete with humans in formulating theories in the social domain without the contextual understanding that being a human automatically affords? Given its dependency on human-generated knowledge (i.e., training data), it is plausible that AI performance in scientific settings will tend to asymptote—at best approaching the performance of top human experts. If this account is correct, AI may offer efficiency gains without improving the quality of social science ideation, especially at the upper end of the distribution, but performance then needs to be directly compared to that of human scientists. One relevant empirical test is whether LLMs can outperform senior topic experts at high-level tasks demanding creativity, such as constructing new social and psychological theories to explain observed correlational patterns.

The present crowdsourced initiative directly compared the performance of 13 senior social scientists, 60 doctoral students from a range of disciplinary backgrounds, and 25 LLMs at advanced theory building challenges concerning academic discourses regarding race and gender inequality. Among the 73 human contributors, 17 were classified as topic experts—11 senior scientists and 6 doctoral students—based on having at least one publication related to racial or gender inequality. A separate panel of scholars with relevant backgrounds generated the constructs of interest and means of measuring them. We sampled a corpus of academic papers from Semantic Scholar. For each paper, the focal outcome of interest was whether its title and abstract discussed racial inequality or gender inequality. An ensemble learning algorithm identified robust empirical predictors of these outcomes using iterative subsample replications[26], and logistic regressions estimated the direction of association for the selected main effects and interactions. The identified empirical relationships served as the machine learning (ML) based empirical benchmarks for the subsequent prediction and theorizing tasks. Crowds of both human scholars and generative AI models attempted to develop *a priori* theoretical models to predict empirical relationships that the ML procedures would uncover, and then revised their theoretical explanations in light of the observed ML results after-the-fact. The survey procedure is presented in Extended Data Fig. 1.

We note that the ML benchmark does not reveal causal relationships, the sample is large but far from comprehensive, and a theory may be scientifically valuable despite disagreement with this benchmark. However, theories can also be judged on the extent to which they anticipated the patterns of robust correlations found in data and offered a coherent causal process that could plausibly generate these patterns. This is one of the standards to which creative or "abductive" insight in theorizing in the social sciences is routinely held[26–30].

This large-scale crowd approach allows for identifying the comparative advantages, if any, of human scientists and artificial intelligences at theory building in the social sciences. We examined not just specific hypotheses but fully articulated theories of a novel phenomenon. We assessed the objective structure and complexity of the theoretical models, in terms of number of paths and latent variables, as well as their subjective quality in the eyes of independent raters. Critically, theorists' predictions about expected correlational patterns and the associated *a priori* theories were put to empirical test, by examining their alignment with realized empirical results[31,32]. This allowed for documenting theory revision in response to new evidence from empirical outcomes, that is, the willingness of human scientists and artificial intelligences to update initial theories in light of potentially contrary findings. We tether human and AI-generated scientific ideas to real empirical outcomes, with a pre–post design that further captures reactions to predictions turning out to be inconsistent with the held-out empirical benchmark. At a bird's-eye level, we examined the overall distributions of human and AI-generated ideas, as well as their respective abilities to form "wise" crowds in the aggregate[33,34].

Three key differentiators stand out among many when comparing human scientists and artificial intelligences engaged in scientific inquiry: processing capacity, emotional detachment, and intellectual diversity. Generative LLMs can combine information from vast pretraining corpora that support broad generalization[35], parallel attention mechanisms[36], and multimodal architectures with million-token context windows[37], together enabling them to operate over information at a scale far beyond unaided human cognition. In contrast, the typical human can hold only a limited number of items in working memory, and long-term memory is subject to decay and confabulation[38–40]. More generally, the human mind satisfices rather than optimizes, defaulting towards heuristic shortcuts to avoid having to fully process complex situations[41]. This points to a possible AI advantage in reasoning over scientific complexity, as reflected in AIs being more likely to develop theoretical models involving numerous causal paths and latent variables, and more likely to integrate conflicting new evidence.

On the latter point, artificial intelligences could further make more dispassionate and detached investigators. Although scientific theories can be extremely useful, they can also obstruct scientific process[42] when confirmation bias[43] sets in and new information is too readily interpreted in light of prior expectations. Daniel Kahneman[44] referred to this as theory-induced blindness: "once you have accepted a theory and used it as a tool in your thinking, it is extraordinarily difficult to notice its flaws." Although LLMs are already known to exhibit some human-like information processing bias, including overconfidence[45], cognitive dissonance[46], and cultural stereotyping[47,48], they could still be relatively *less* biased in favor of their prior theories than humans[45], providing a potential benefit to scientific research projects.

A key collective strength of human populations may lie in cognitive diversity—i.e., divergent ways of thinking about the same problem[49]. Even if many individual humans endorse erroneous theories, these errors tend to be randomly distributed, in some contexts leading to wise crowds when predictions are aggregated[34]. Further, the tendency for some individual humans to gravitate towards beliefs that diverge greatly from the mainstream may introduce precious atypical insights into the collective discussion. By comparison, past work suggests that LLMs tend to produce more homogeneous responses than human populations[50,51]. In the domains of fiction writing and product design, AI use has been shown to raise individual creativity but also reduce collective diversity[52–54]. In terms of overall structure of the provided theoretical explanations, this predicts a sparse core + long tail structure for human-generated scientific theories of the same phenomenon, and a concentrated core for AIs. We also examined whether human researchers would benefit more than artificial intelligences from aggregation in terms of their predictive accuracy.

Most broadly, our goal was to begin to identify the comparative advantages of artificial intelligences and human minds for theory building in the social sciences, specifically their ability to predict correlational empirical patterns and the differences in their constructed causal explanations that can plausibly account for these patterns, with an eye towards a future of scientific inquiry in which human–AI collaboration prevails as the new norm[8,55–57].

# Results

### R.1 Humans and generative AI exhibit complementary strengths in making predictions about correlational patterns

**GenAI matched or exceeded human predictions at the individual level.** We first examined how accurately human contributors and generative AI systems identified the most important main effects and second-order interactions, as well as the directions of those associations, for predicting the presence of racial and gender inequality discourses across a corpus of published peer-reviewed publications (Fig. 1 and Extended Data Table 1). Predictive accuracy was evaluated against the ML benchmarks derived using random-forest models to identify the most predictive features and logistic regression to estimate the direction of their associations (Extended Data Fig. 2 and Extended Data Fig. 3). These correlational patterns were robust to sampling error through a variety of re-and sub-sampling procedures standard in machine learning. These ML results identify the strongest associations expected to replicate out of sample, which are precisely the patterns our theorists were asked to predict and explain, though they do not necessarily represent causal relationships. Our theorists were aware that the data were observational, based on a corpus of academic papers, and not generated through randomization. Their task was therefore to anticipate empirical associations in the data, while providing causal theories that could explain those associations.

Predictive accuracy scoring ranged from −1 (complete opposition) to 1 (perfect alignment), with 0 indicating no directional alignment. For main-effect predictions, both human contributors and generative AI systems performed well above the zero baseline corresponding to the random benchmark (Fig. 1). GenAI closely matched humans on predicting the main effects for racial inequality discourses with no significant difference (0.416 vs. 0.411, $P = 0.820$) and outperformed them on the gender-inequality discourse task (0.624 vs. 0.578, $P = 0.026$; Extended Data Table 1).

Second-order interactions posed a greater combinatorial challenge than main-effect tasks because contributors had to identify combinations among 78 possible feature pairs and anticipate the direction of their joint effects. GenAI systems showed their largest individual-level advantage on racial inequality interactions (0.400 vs. 0.128, $P < 0.001$). By contrast, both groups performed near chance on the gender-inequality interaction task. Thus, GenAI was already competitive with, and in some settings superior to, individual human scientists, with a large advantage on the racial inequality interaction prediction task but not the equivalent task for gender inequality.

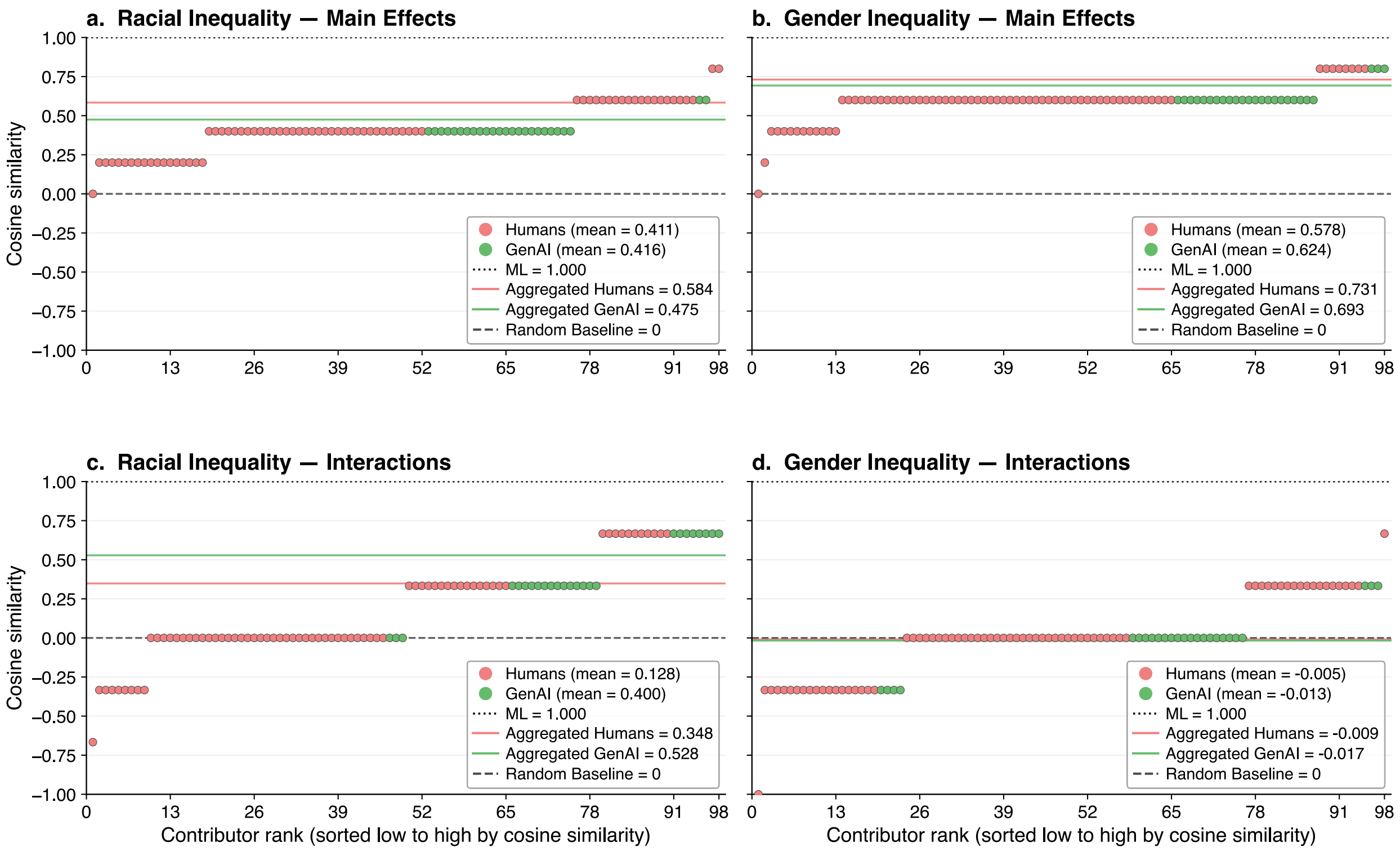

**Fig. 1 | Prediction accuracy across tasks and model specifications. a,b,** Cosine similarity between predictions and the ML benchmark for main effects, for racial and gender inequality, respectively. **c,d,** Cosine similarity between predictions and the ML benchmark for second-order interactions, for racial and gender inequality, respectively. Points represent individual contributors, ordered by predictive accuracy and colored by group. Solid horizontal lines indicate the accuracy of the aggregated human and GenAI predictions, dashed lines indicate the zero baseline corresponding to the random benchmark, and dotted lines indicate the ML benchmark (= 1).

**Human collectives drew greater value from diversity at the aggregate level.** At the collective level, predictions aggregated across all human contributors ($N = 73$) were significantly more accurate than those aggregated across all GenAI systems ($N = 25$) on the racial inequality main-effect task (0.584 vs. 0.475, $P = 0.023$) and directionally but non-significantly more accurate on the gender-inequality main-effect task (0.731 vs. 0.693, $P = 0.206$; Extended Data Table 1). In contrast, GenAI exhibited a non-significant directional aggregated advantage on the racial-inequality interaction task (0.528 vs. 0.348, $P = 0.174$; Fig. 1). The human aggregated advantage for main effects persisted across equal-sized resampled crowds of 2–25 contributors, indicating that it was not driven by the larger number of human contributors (Fig. 2a,b).

To quantify the incremental value created by aggregation, we defined *aggregation gain* as the accuracy of the aggregated predictions minus the mean accuracy of the individual theorists within each crowd. Across equal-sized resamples, human crowds generally achieved larger aggregation gains than GenAI crowds (Fig. 2e-h). We found that greater prediction heterogeneity, measured as prediction diversity and error diversity, was associated with larger aggregation gains (Extended Data Table 2). Human crowds exhibited significantly greater prediction heterogeneity than GenAI crowds on both measures across four task-by-effect comparisons (all eight $P$s < 0.001) and were, on average, 86% more diverse, providing a compositional basis for their larger gains from aggregation.

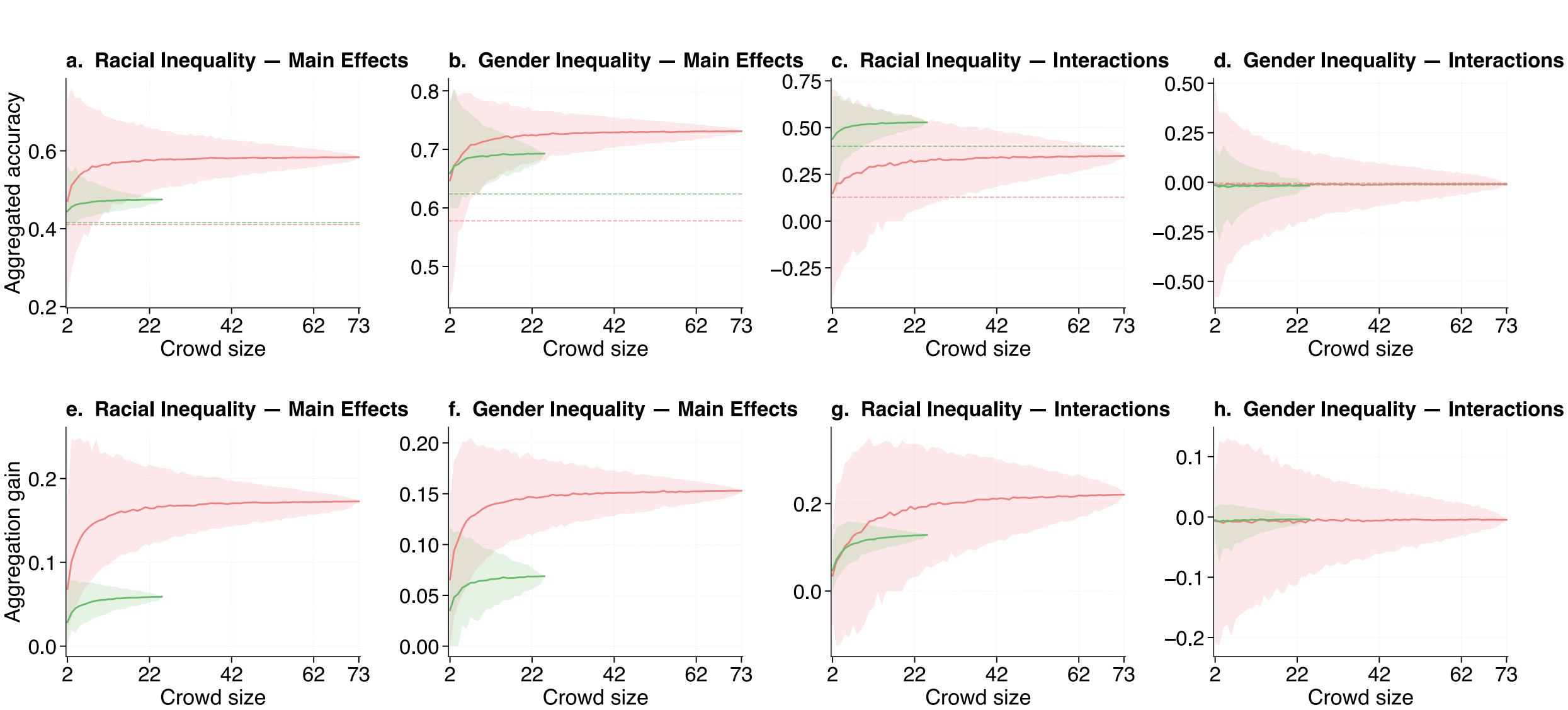

**Fig. 2 | Aggregated accuracy and aggregation gain across equal-sized human and GenAI crowds.** Top row (**a–d**): aggregated accuracy for main-effect (**a, b**) and second-order interaction (**c, d**) predictions on racial and gender inequality, respectively. Bottom row (**e–h**): corresponding aggregation gains. Curves are means over 1,000 without-replacement resamples of crowds; bands are 95% resampling intervals. Dashed lines in a–d show mean individual accuracy.

**Two common error patterns: domain-specific theories and wrong direction.** Predictive errors revealed two common patterns across humans and GenAI. First, they tended to make conceptual silo errors, failing to realize that gender-related variables were predictive of race discourses and vice versa. In the racial-inequality main-effect task, only 18 of 98 contributors accurately selected *female*, the estimated share of female authors in the focal paper (Extended Data Fig. 4a), despite its ranking among the strongest ML predictors (Extended Data Fig. 2a). A similar pattern emerged in the gender-inequality main-effect task, where only 6 of 98 contributors selected *Asian*, the estimated share of Asian authors in the focal paper (Extended Data Fig. 4b), despite its ranking among the strongest ML predictors (Extended Data Fig. 3a). Conversely, theorists favored salient, domain-specific relationships even when they were not supported by the benchmark. For example, *black * social_science*, the interaction between the estimated share of Black authors and whether the paper was published in a social science journal, was most frequently selected for the racial-inequality interaction task (Extended Data Fig. 4c), although it did not rank among the strongest ML interactions.

Second, theorists sometimes identified an important relationship but assigned it the wrong sign. In the interaction tasks, many contributors identified *social_science* ∗ *female* as important but assigned it the wrong direction. The ML benchmark identified the direction of this interaction as negative, whereas 82% of selectors in the racial task and 100% selectors in the gender task assigned positive signs (Extended Data Fig. 4c,d). This suggests that predictive failures arose not only from difficulty identifying relevant predictors, but also from difficulty anticipating when plausible relationships operated differently from shared expectations (e.g., two linked variables substituting for one another rather than amplifying each other's effects). These error patterns are descriptive and based on an exploratory internal analysis, supplying hypotheses for future confirmatory work.

**Scientific seniority and topic expertise were most consequential for the most difficult prediction task.** Although it is intuitive that experienced senior scientists and those who study a given topic themselves should hold a prediction advantage, the prior empirical evidence on this point is mixed—with topic expertise (e.g., measured by prior publications on the topic) and scientific seniority (e.g., measured by job rank) associated with increased accuracy in some but not all studies[32,58–62]. In the present investigation, individual performance was broadly comparable between PhD students and more senior scientists (professors), and between topic experts and non-experts, with regard to predicting main effects (Extended Data Table 1 and Extended Data Table 3). However, scientific seniority and topic expertise did confer some advantage on the more challenging predictions regarding interactions, which required contributors to select from a substantially larger set of candidates than the main-effect tasks. On the racial inequality interaction task, senior scientists exhibited higher directional accuracy than PhD students (0.205 vs. 0.111, $P = 0.347$), and topic experts exhibited higher directional accuracy than non-experts (0.196 vs. 0.107, $P = 0.312$; Extended Data Table 1). On the gender-inequality interaction task, where mean accuracy for average humans and GenAI systems was near chance, topic experts achieved significantly higher accuracy than non-experts (0.118 vs. −0.042, $P = 0.038$, Extended Data Table 1; Expertise coefficient $\beta_2 = 0.217$, $P < 0.05$, Extended Data Table 3) and marginally exceeded GenAI systems (0.118 versus −0.013; $P = 0.084$). This suggests that expertise may matter most when scientific predictions are especially challenging (i.e. typical predictor does poorly). However, such a general conclusion cannot be supported by a single isolated result across a small set of tasks. Future meta-analyses and new pre-registered data collections should provide better powered confirmatory tests of these exploratory patterns.

### R.2 GenAI theories were more complex, whereas human theories were more diverse

**Theory complexity: GenAI-produced theories were structurally more elaborate than human-produced theories.** GenAI causal diagrams contained more causal paths, more latent variables, and longer causal chains than human diagrams, and GenAI textual theoretical explanations contained substantially more words (all $P$s ≤ 0.027; eight of 16 comparisons $P \leq 0.001$; Extended Data Table 4a). Within the human group, differences by academic rank and topic expertise were generally not statistically significant, although senior scientists and topic experts showed a tendency toward *lower* structural complexity on several diagram-based measures (Extended Data Table 4a). Before exposure to the ML evidence, GenAI diagrams typically contained approximately seven paths and five latent variables, whereas human diagrams were relatively simpler, typically containing approximately five paths and two latent variables; GenAI explanations were also roughly two to three times as long as human explanations (approximately 270–300 vs. 120–130 words). Following exposure to the ML evidence, GenAI theories became even more complex: mean path counts increased from 6.64 to 8.28 ($P < 0.001$) and from 6.80 to 8.56 ($P = 0.014$), mean latent-variable counts from 5.16 to 6.72 ($P = 0.004$) and from 5.52 to 7.28 ($P < 0.001$), and mean theory word counts from 273.44 to 320.44 ($P = 0.007$) and from 300.50 to 357.80 ($P = 0.007$), for racial and gender inequality, respectively. In contrast, human theories showed no significant pre–post changes in these complexity measures.

**Theory diversity: Human theories were more diverse.** We quantified the theory diversity of the provided textual theoretical explanations along three dimensions: semantic coverage, dispersion, and distributional structure. Human theories were more semantically diverse across all four tasks and in both the pre- and post-ML results phases. First, two-dimensional PCA projections of the 3,072-dimensional explanation embeddings showed visually that both human groups—PhD students ($N = 60$) and senior scientists ($N = 13$)—spanned a broader region of semantic space than GenAI systems ($N = 25$; Fig. 3). Second, quantitative analyses in the 3,072-dimensional embedding space showed significantly greater within-group dispersion among human explanations than GenAI explanations, as measured by both the cosine distance to the group centroid and the within-group pairwise cosine distances (all $P$s < 0.001; Fig. 4a-d). On average, the distances of human explanations from their respective group centroids were approximately 2.4 times those of GenAI explanations (Fig. 4a,c), and the pairwise distances between human explanations were approximately 2.2 times those between GenAI explanations (Fig. 4b,d). No significant difference in within-group dispersion was observed between PhD students and senior scientists (Fig. 4e-h). Third, HDBSCAN clustering revealed distinct core–tail structures. We defined explanations assigned by HDBSCAN to a semantic cluster as belonging to the core, and explanations labeled as noise points as belonging to the tail. Across tasks, effects and phases, only 9.6–31.5% of human explanations belonged to the core, compared with 92–100% of GenAI explanations (Extended Data Fig. 5a). A sensitivity analysis varying the minimum cluster size from 2 to 10 showed that the qualitative human–GenAI differences in core fraction were robust to this choice (Extended Data Fig. 5b). Human theories therefore formed a relatively sparse core surrounded by a long tail of distinct explanations, whereas GenAI theories were highly concentrated within a single dense core in each condition, when arrayed in a common semantic space.

**Revision after exposure to evidence reduced human theory diversity, but human theories remained more diverse.** Exposure to the ML evidence reduced the semantic diversity of human explanations across all four tasks, but did not eliminate their greater diversity relative to GenAI. Human theories showed significantly lower within-group semantic dispersion after exposure to the ML evidence across all four tasks, with cosine distance to the group centroid declining by an average of 12.6% (Fig. 4c) and mean pairwise cosine distance by 11.5% (Fig. 4d). In contrast, GenAI dispersion remained largely unchanged before and after exposure to the ML evidence. The share of human explanations assigned to the HDBSCAN clustering core also increased from 9.6–17.8% before exposure to 21.9–31.5% afterwards (Extended Data Fig. 5a). Human explanations nevertheless remained more semantically dispersed than GenAI theories and retained a more pronounced long-tail structure after exposure to ML evidence, with 68.5–78.1% remaining outside the semantic core.

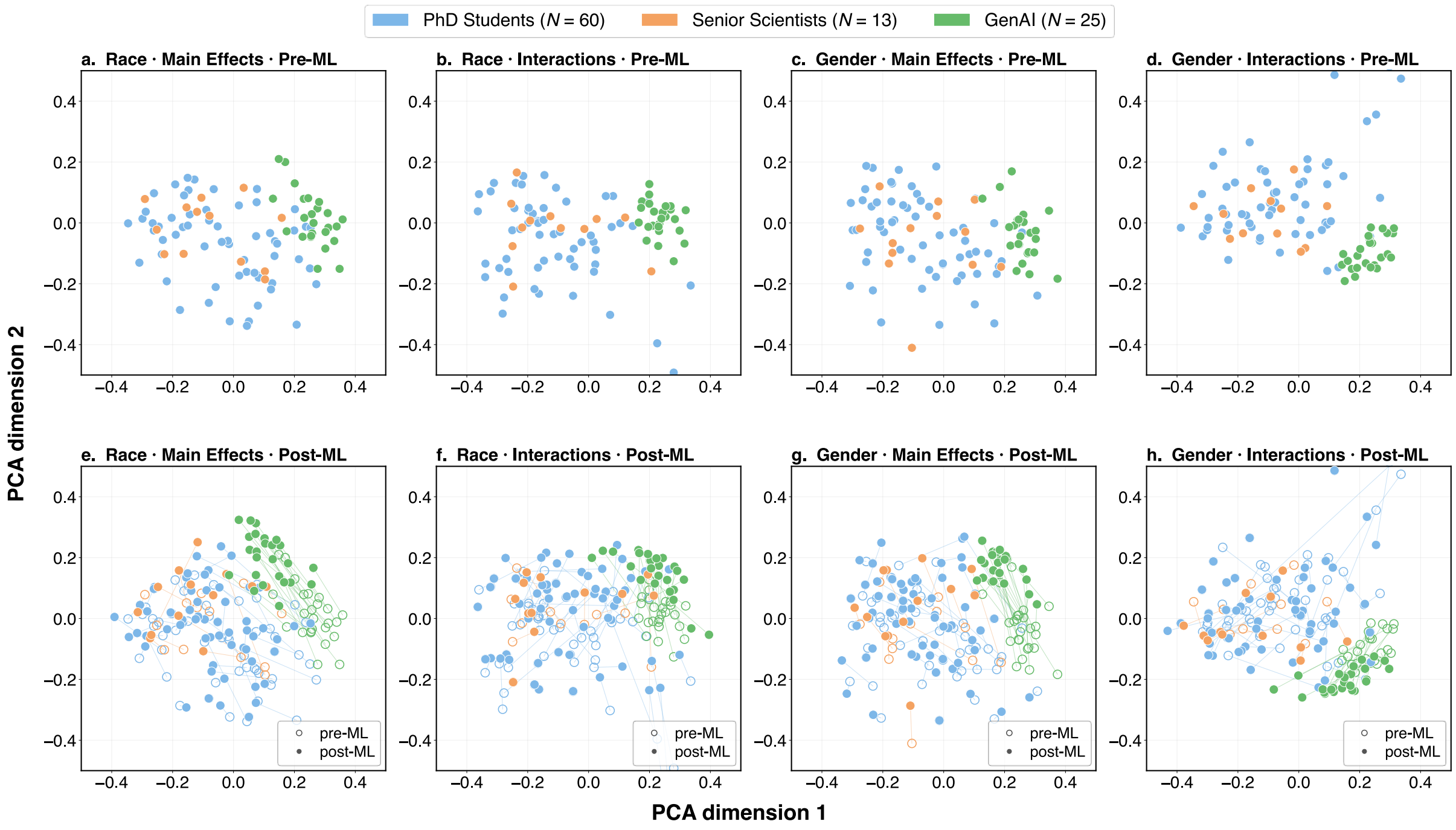

**Fig. 3 | Semantic distributions of theoretical explanations across PhD students, senior scientists and GenAI systems**. Two-dimensional PCA projections of the 3,072-dimensional embeddings of theoretical explanations across the racial- and gender-inequality tasks in the pre- and post-ML phases. Each point represents one theoretical explanation, with colors indicating contributor group. In the post-ML panels (**e-h**), open circles show pre-ML positions, filled circles show post-ML positions, and lines connect the same contributor's pre→post trajectory.

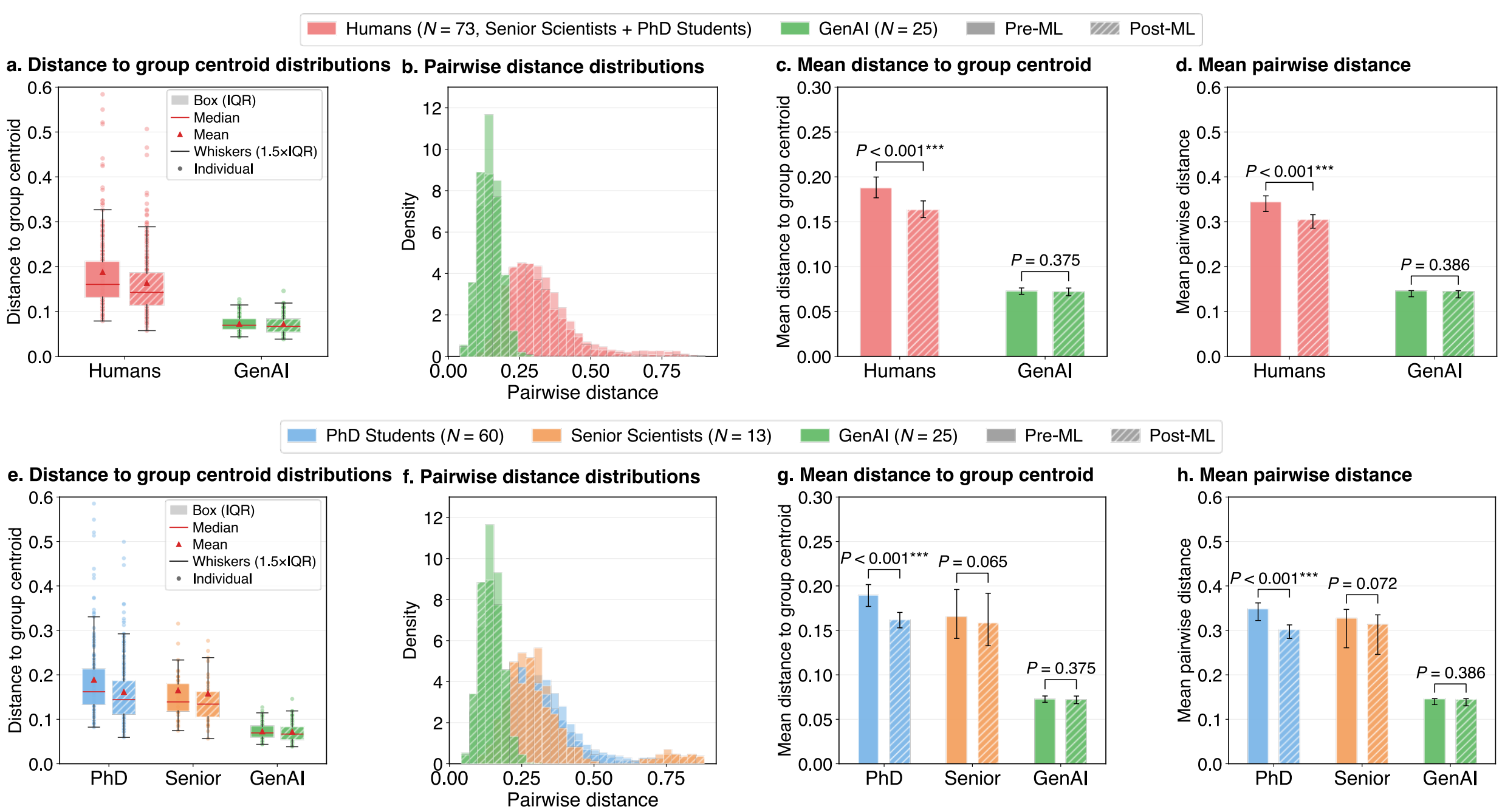

**Fig. 4 | Within-group semantic dispersion before and after exposure to ML evidence, pooled across the four task–effect combinations. a,e**, Distributions of cosine distances in the 3,072-dimensional embedding space between individual theoretical explanations and the centroid of their respective contributor group. **b,f**, Density-normalized distributions of pairwise cosine distances in the embedding space between explanations from the same contributor group. **c,g**, Mean cosine distance to the corresponding group centroid before and after exposure to ML evidence. **d,h**, Mean pairwise cosine distance within each contributor group before and after exposure. **a–d** compare all human contributors with GenAI systems, whereas **e–h** show PhD students and senior scientists separately. In **c,d,g,h**, bars show means and error bars indicate 95% confidence intervals. Solid and

hatched elements indicate pre-ML and post-ML explanations, respectively. Higher values indicate greater within-group semantic dispersion. $P$ values are from one-sided paired tests of whether post-ML dispersion is lower than pre-ML dispersion: paired $t$-tests in **c** and **g**, and paired permutation tests with 5,000 permutations in **d** and **h**. *** indicates $P < 0.001$.

### R.3 Human updating was more sensitive to prior errors, whereas GenAI revised more extensively

**GenAIs revised their theories more extensively in response to new evidence.** We measured the extent of theory revision as the cosine distance between each contributor's pre- and post-ML explanation embeddings, with larger values indicating greater semantic change. On average, GenAI systems exhibited 42% larger pre- to post-ML semantic revisions than human contributors (0.145 vs. 0.102; $P < 0.001$; Fig. 5a). These larger revisions, however, did not translate into greater diversity across GenAI systems. After exposure to the evidence, their explanations showed little overall change in within-group semantic dispersion (Fig. 4), remaining concentrated within a relatively narrow region of semantic space (Fig. 3e–h), and were assigned entirely within a single dense semantic core in each condition (Extended Data Fig. 5a). GenAI systems therefore changed their explanations more extensively while continuing to produce a homogeneous set of theories.

**Human updating was sensitive to prior prediction errors.** The two groups differed not only in how much they revised, but also in *when* they revised. Among human contributors, lower pre-ML predictive accuracy was associated with larger subsequent revisions (OLS $\beta = -0.098$; Spearman $\rho = -0.31$; both $P$s $< 0.001$; Fig. 5b). Thus, humans changed their theories more when their initial predictions were inconsistent with the ML evidence. No comparable relationship was observed among GenAI systems (OLS $\beta = 0.008$; Spearman $\rho = 0.08$; both $P$s $\geq 0.05$; Fig. 5c). This difference was confirmed in a mixed-effects model controlling for task fixed effects (Fig. 5d). Among humans, revision increased with prior prediction error (slope for humans = 0.105, $P < 0.001$), whereas no comparable association was observed among GenAI systems. Also, the interaction was significant ($\beta_3 = -0.091$, $P = 0.003$), confirming that the error–revision association was stronger for humans than for GenAI systems. To the extent that the theorist's role is to refine and not merely formulate theory, it appears that AIs are more active refiners, but human theorists are more sensitive refiners.

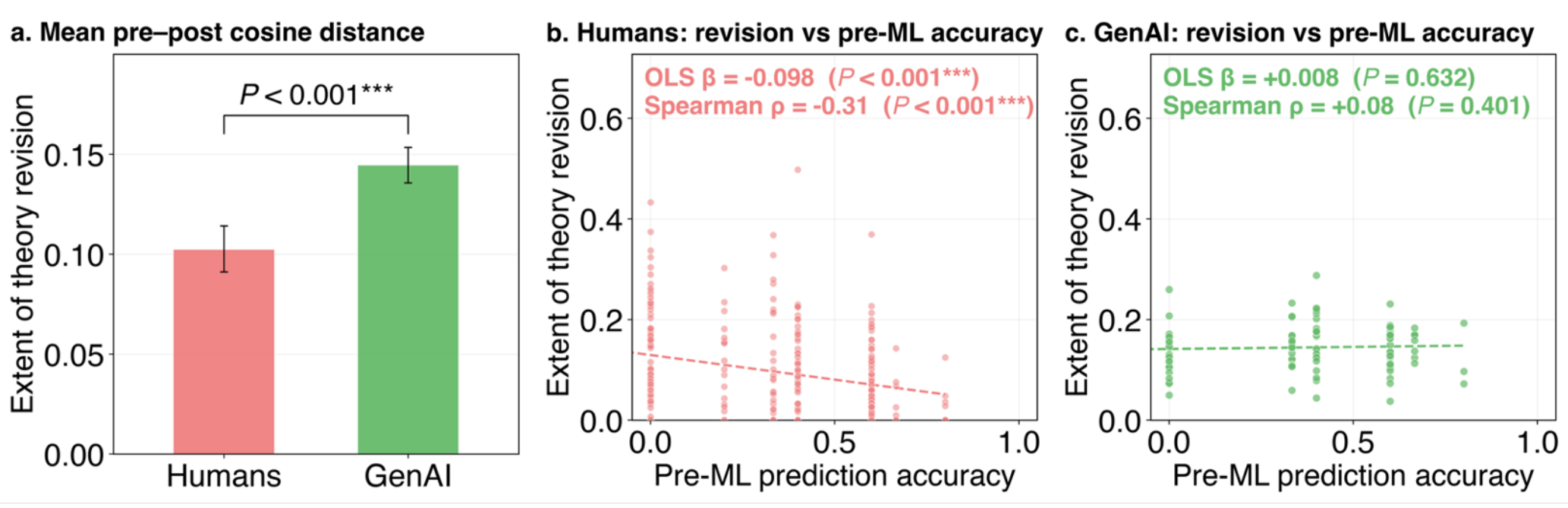

**d Mixed-effects model of theory revision**

| | Intercept ($\beta_0$) | Prior error ($\beta_1$) | GenAI ($\beta_2$) | Prior×GenAI ($\beta_3$) | Slope (Humans) | Slope (GenAI) |
|---|---|---|---|---|---|---|
| Coef. | 0.028* | 0.105*** | 0.109*** | −0.091** | 0.105*** | 0.015 |
| 95% CI | [0.003, 0.053] | [0.069, 0.142] | [0.061, 0.156] | [−0.150, −0.031] | [0.069, 0.142] | [−0.047, 0.077] |

**Fig. 5 | Extent of theory revision and its association with prior prediction accuracy. a,** Mean cosine distance between pre- and post-ML explanation embeddings for human contributors and GenAI systems. Bars show group means, and error bars indicate bootstrap 95% confidence intervals obtained by resampling observations. The significance marker reports a two-sided Welch's $t$-test comparing the two groups**. b,c,** Association between pre-ML prediction accuracy and the extent of theory revision for human contributors and GenAI systems, respectively. Each point represents one contributor's observation for one of the four tasks; observations are pooled across tasks. Dashed lines show ordinary least-squares fits. The OLS slope and Spearman rank correlation are reported; significance markers refer to the corresponding slope and rank-correlation tests. **d,** Mixed-effects model of theory revision. Let $j$ index contributors and $t$ index task-by-effect conditions. $\text{Theory Revision}_{jt}$ is the cosine distance between contributor $j$'s pre-ML and post-ML theory embeddings in condition $t$, and $\text{Prior Error}_{jt} = 1 - \text{pre-ML accuracy}_{jt}$. $\text{GenAI}_j = 1$ for GenAI contributors and 0 for humans (reference). The model is $\text{Theory Revision}_{jt} = \beta_0 + \beta_1 \text{Prior Error}_{jt} + \beta_2 \text{GenAI}_j + \beta_3 \left(\text{Prior Error}_{jt} \times \text{GenAI}_j\right) + \delta_t + u_j + \varepsilon_{jt}$, where $\delta_t$ are task-by-effect fixed effects and $u_j$ is a contributor-level random intercept. Thus $\beta_1$ is the Human slope and $\beta_1 + \beta_3$ is the GenAI slope. Cells report coefficients with 95% confidence intervals, with *, **, and *** indicating $P < 0.05$, $P < 0.01$, and $P < 0.001$, respectively.

## R.4 Human and LLM independent raters agreed in rating GenAI theories more highly

**Human and LLM evaluators showed strong agreement in their assessments of theory quality.** The LLM evaluator rated the full set of theories, whereas two independent senior academics, one with expertise in human–AI interaction and the other on race and gender inequality, rated a random subset of the theory corpus. All evaluators were blinded to theory provenance, including contributor type and whether the theory was produced before or after exposure to the ML evidence. Ratings were strongly correlated (Pearson $r = 0.82$, Spearman $\rho = 0.82$, both $P$s < 0.001; Extended Data Fig. 6a), indicating substantial agreement in both scores and rank ordering across human and LLM independent ratings.

**GenAI theories received substantially higher quality ratings than human theories across both human and LLM evaluation.** GenAI theories were rated significantly higher than those produced by PhD students, senior scientists, and topic experts (all $P$s < 0.001; Extended Data Fig. 6b,c). Differences in assessed theory quality among human contributors (PhD students vs. senior scientists; topic experts vs. non-experts) were not significant using the smaller sample of human independent ratings, whereas the LLM evaluation across the full sample of theories showed significant advantages for senior scientists and topic experts (all $P$s < 0.05 for both pre- and post-ML results; Extended Data Fig. 6b). Following ML results exposure, ratings of human-authored theories increased significantly in both the ratings from human evaluators on a subset of

theories ($P = 0.019$) and the LLM evaluation across the full sample of theories ($P = 0.020$), whereas ratings of GenAI theories remained largely unchanged.

### R.5 Theory complexity is associated with perceived quality but not predictive accuracy

**Theory complexity is associated with subjective quality but not the accuracy of scientific predictions.** Across all four complexity measures, the estimated association between theory complexity and prediction accuracy was small and generally statistically non-significant for both human and GenAI contributors: human slopes ranged from −0.013 to 0.020, whereas GenAI slopes ranged from −0.020 to 0.002, with all 95% confidence intervals spanning zero (Extended Data Table 4b). The complexity–GenAI interaction terms were likewise non-significant across complexity measures, except for maximum path length ($\beta_3 = -0.04$, $P = 0.025$). In contrast, theory complexity was significantly associated with *perceived* theory quality. In particular, a greater number of causal paths predicted significantly higher LLM-assessed quality ratings for both humans and GenAI theory contributors (human slope = 0.531, $P = 0.001$; GenAI slope = 0.231, $P = 0.002$; Extended Data Table 4c). Theory word count showed the same pattern (human slope = 2.154, $P < 0.001$; GenAI slope = 0.220, $P < 0.001$; $\beta_3 = -1.935$, $P < 0.001$). Greater maximum path length and a larger number of latent variables were each associated with higher quality ratings for GenAI theories but not for human theories (Extended Data Table 4c). Thus, among GenAI theories, greater structural complexity was not associated with improved prediction accuracy but was associated with higher *perceived* theory quality.

**Human theorists are more parsimonious.** At mean complexity, GenAI systems received substantially higher quality ratings than human contributors across all four complexity measures ($\beta_2$; Extended Data Table 4c); the corresponding GenAI accuracy advantage was significant for theory word count ($\beta_2 = 0.086$, $P < 0.001$) but not for the three diagram measures (Extended Data Table 4b). Yet humans achieved significantly higher prediction accuracy per unit of complexity on all four measures ($P$s $\leq 0.006$; Extended Data Table 4d), consistent with the finding that GenAI's greater structural complexity was not accompanied by proportionately greater predictive performance. Assessed quality per unit of complexity did not differ significantly for the three diagram measures, but humans obtained higher quality per word (0.051 vs. 0.032, $P < 0.001$; Extended Data Table 4d).

## Discussion

The present crowdsourced initiative competed scores of scientists and generative LLMs at building empirically grounded theories about the incidence of discourses regarding race and gender inequality. Provided with descriptions of the sets of independent and dependent variables, as well as the statistical analyses involved, human scientists and GenAI systems attempted to develop *a priori* theoretical models and predict relevant empirical relationships. Then, after learning about the realized findings, they provided *post-hoc* theoretical explanations for the observed patterns, revising or defending their initial theoretical models. A crowd approach allowed us to compare human scientists and artificial intelligences at both the individual and collective levels.

### *Possible AI advantages in theoretical reasoning over complexity*

Remarkably, individual generative LLMs matched, and in some cases outperformed, the average human scientist in our sample at a number of these advanced theory building challenges. Theoretical models developed by AIs were on average preferred by independent expert raters blind to their source. GenAI theories were also more extensively elaborated and involved more causal paths and latent variables. This complexity was to some extent ornamental in that although impressive to evaluators, both human and machine, it was not accuracy-enhancing. Thus, one might prefer the human theories on the grounds of parsimony and efficiency.
And yet, AI-generated theories were not merely impressive sounding nonsense[23,24], in that the predictions generated from AI theories aligned more closely with the realized empirical patterns than did those generated from human theories on two of the four tasks. Indeed, GenAI substantially outperformed humans at predicting racial inequality interactions, achieving an accuracy score approximately three times that of humans.

Human expertise made no difference on the comparatively easy main-effect tasks and was more visible on the harder interaction predictions, where anticipating how variables combine requires a genuine causal model rather than surface pattern-matching. On the gender-inequality interaction task, where average human and GenAI performance were both near chance, topic experts and senior scientists performed well above it, directionally outperforming GenAI, suggesting that human expertise may become most visible on especially difficult prediction problems. Yet at the same time, four tasks—race and gender main effects and interactions—cannot establish a general complexity gradient. Further, although the largest AI advantage emerged in one of the two interaction tasks, this was not consistent across both tasks. We are therefore unable to draw the strong inference that GenAI holds an advantage for predicting more complex empirical patterns, or that experienced scientists and topic experts are able to match or exceed GenAI for some especially difficult-to-predict outcomes. We leave empirical confirmation of these possible moderating factors to future confirmatory investigations.

The generative LLMs were comparatively more responsive to new information, and quicker to abandon their initially elaborated theories when presented with new data. Despite the fact that their theories were considerably *less* likely to enjoy empirical support, human scientists were significantly *more* likely to stick with their original theoretical model or something close to it. Interestingly, human and artificial intelligences further differed in *when* they updated their theories. We observe that humans appear to exhibit an error-sensitive updating mechanism: respondents with less accurate pre-ML predictions tend to make larger semantic revisions after seeing the ML output. In contrast, we do not observe such calibration in GenAI systems—the magnitude of their theoretical revisions was largely unrelated to the (in)accuracy of their initial predictions. Although speculative, GenAI systems may have approached the post-ML task as a new generation problem conditioned on expanded context rather than as an incremental revision of a prior theory. Thus, their larger semantic shifts may reflect contextual regeneration rather than belief updating calibrated to prior error. Although each LLM had its pre-ML response in its context when asked to provide a theory update, they may not maintain an explicit belief state at all, but rather follow a stateless, prompt-driven revision process.

A complexity-moderated story emerges from the above meta-scientific results. Due to the processing capacity limitations of the human mind, even trained scientists may face significant challenges enumerating theories with numerous causal pathways and variables and integrating conflicting evidence that challenges established beliefs. On a positive framing, human theorists were more parsimonious, a virtue in scientific theorizing so long as explanatory power is not sacrificed. Indeed, human theorists were more predictively efficient, exhibiting higher prediction accuracy-to-complexity ratios across four distinct measures of complexity. The limits of human cognition may act as a regularization mechanism that prevents complexity if it does not enhance accuracy.

As noted earlier, some empirical evidence indicates LLMs exhibit a number of human-like judgmental tendencies, including cognitive dissonance[46]. At least in this setting, their ability to discard prior theories based on new evidence and thus avoid theory-induced blindness[42,44] compared favorably to human scientists. Given that our contributors had only recently formulated and committed to their *a priori* theoretical models, we see this as a conservative test of an AI advantage in this regard. Active scientists would presumably have much more difficulty letting go of published theories they had invested years of their careers into and were both professionally identified with and emotionally invested in. And yet, it is not clear if AI should be characterized as comparatively open-minded, as this may anthropomorphize a text-generation process.

***The benefits of human intellectual diversity***

The possible downside risks of displacing human theoreticians with artificial intelligences were visible primarily in the aggregate. Human theories were more semantically diverse in terms of breadth, dispersion, and distributional structure. Specifically, the distribution of human-generated theories exhibited a sparse core with a long tail. In contrast, AI theories were characterized by a concentrated core structure, lacking the outliers as well as overall heterogeneity of human ideas. Of course, semantic distinctiveness alone does not establish scientific value. Not in spite of, but partly because of their more heterogeneous predictions and errors, human researchers' predictions benefited more from aggregation. In the case of predicting main effects, where the AI advantage at an individual level was small to begin with, this wisdom of the crowd effect[33,34] in some cases led to an overall human advantage. Specifically, the human researchers' collective predictions more accurately predicted simple main effects on the race task relative to the aggregated predictions of the generative LLMs.

The present findings underscore widespread concerns about the macro-level consequences of the similarities of generative LLMs to one another[50–54]. Even if they increase the productivity of individual scientists or raise the quality of the modal scientific article, the increasing prevalence of AIs may ultimately homogenize collective output. That is, AI dependence may reduce the diversity of the paths of inquiry pursued, as well as the diversity of novel and useful theories subsequently introduced into the literature. At a system level, science embraces failure and risk, accepting innumerable unsupported but interesting ideas as necessary to achieve the rare breakthrough contribution. A future in which even high-level scientific tasks are frequently delegated to generative LLMs, which further carry out a host of parallel investigations entirely autonomously[11,12], could prove an era of increased productivity coupled with a drift towards a scientific monoculture. In our setting, human crowds gained more from aggregation than AI crowds because they were far more diverse, and this advantage held even at equal crowd size. One possible explanation is that individual LLMs already aggregate large and substantially overlapping bodies of knowledge through pretraining, such that broader coverage of shared knowledge may leave less room for between-model diversity and make each additional LLM contribute less independent information than an additional human scientist does to a human collective. The durable human strength thus lies in the diversity of the scientific community rather than in any individual theorist—a diversity that substituting homogeneous AI contributions for those of human scientists could erode.

At the same time, diversity in science is not always equally beneficial. In some domains there are single theoretical representations of the phenomena that outperform others in terms of explanatory power; in others, there may be many possible theories with comparable predictive power, and it is difficult to locate the global peak. Cognitive diversity in initial beliefs about theory may matter most on such multi-peaked theoretical landscapes, where local optima are separated by valleys of lower fitness[63,64]. In such settings, heterogeneous theory contributors may keep exploring alternatives in different regions, whereas homogeneous contributors may tend to converge prematurely on a local peak. Some science problems, especially in formal and natural sciences such as mathematics, physics, and chemistry, resemble a single-peaked landscape with one correct answer. However, social science problems may often be multi-peaked, with more than one high-quality answer or valuable perspective, calling for greater intellectual heterogeneity—and consequently perhaps more human involvement as well. Future work should examine empirically the value of diversity in theory tasks situated in multi-peaked theoretical landscapes.

***Limitations and future directions***

Moving from comparison to organizational design, the most productive applied question is not "Who is superior?" but rather "How should scientific collaborations balance the complementary strengths and weaknesses of human scientific experts and artificial intelligences?" The present patterns suggest that expertise might be allocated in a discriminating manner rather than uniformly staffed. The value of domain expertise appeared mainly on the most challenging prediction task. Effective human–AI teams might consider routing predictions of simple relationships to crowds of non-experts and GenAI, who exhibit complementary strengths, and concentrate scarce expert attention on the more challenging problems that neither the crowd of non-experts nor GenAI reliably solves. And yet, the tasks of theorizing on simple and complex data patterns are not always separable—for example, explaining a complex interaction pattern may build on the mechanism proposed for the underlying simple effects. This suggests experimenting with iterative human-AI crowd collaborations in which experts participate throughout in developing and evaluating explanations that progressively increase in complexity.

Hybrid scientific teams could be further designed to exploit and combine the best elements from human theorists and AIs: human tail exploration and error-sensitive theory revision, with GenAI's ability to handle complexity and willingness to abandon prior theories. Given the high utility but homogeneous outputs of GenAI, team assembly might aim to maximize cognitive diversity to enhance the benefits of aggregating scientific predictions and judgments within the group of collaborators. The present results suggest a testable "one AI, many humans" prediction for future research: a team of several human scientists and one strong GenAI may outperform all-human, all-AI, or "one human, many AIs" teams for scientific collaboration.
To incentivize diversity, surveys investigating team design could reward members through an “accuracy-weighted inverse beauty contest,” with scores that reward not just accuracy of theoretical predictions, but also deviation of contributed theories from the group centroid in the semantic space. This should lead theory builders to strive to be different from what they anticipate the field’s centroid will be. Future research should also examine whether human–AI collaboration preserves the respective strengths of human and AI contributions in theory building, and whether its benefits outweigh the coordination costs of integrating such different approaches.

Alternative explanations for our results are important to consider. Some of our human contributors may have under-invested effort in their theories and scientific predictions, such that apparent AI advantages partly reflect lower human effort rather than genuinely stronger AI performance. The available evidence indicates that crowd data collections are, if anything, associated with lower error rates than small team collaborations, consistent with the idea that crowdsourced projects select for the already highly motivated contributors (see Uhlmann et al.[65] for a review). Although, as noted in the SI, we did identify two outlier cases where respondents clicked through the survey in less than 20 minutes, these were removed from all the analyses reported here. After accounting for potential pauses in survey completion by excluding unusually long completion time, the remaining scientists spent an average of 2.69 hours on the survey (median = 2.15 hours; $SD$ = 2.34), suggesting meaningful expenditures of effort. Future meta-scientific comparisons of artificial intelligences and human scientists should use incentivized surveys, ideally manipulating the payoffs for predictive accuracy and independent quality ratings to directly address the motivation issue.

Despite its scale, this crowdsourced initiative ultimately represents an $N = 1$ case study of human and AI comparative advantages in scientific theory building. Before drawing strong inferences[66], many more such meta-scientific studies employing diverse paradigms to address a wide range of scientific questions are needed. The specific challenge involved here—anticipating and explaining patterns in large-scale text data based on a fixed set of available variables—could favour LLMs[32,67]. Human scientists were required to fit theories into a structured format that may have limited their natural advantages and creativity, while AI models may be particularly skilled at producing content in pre-specified formats. We find that GenAIs often match and sometimes exceed humans at theory building challenges in this task environment and using these models, prompts, and benchmarks, but this may not be true generally. AIs may struggle at theory-building tasks that involve inferential leaps as opposed to recombining existing knowledge[68], and engage in violations of scientific integrity by failing to properly credit the prior scholarship they draw on[69]. Follow-up investigations should directly compare the ability of crowds of AIs and human scientists to ideate and design novel experimental studies and empirical tests of proposed theoretical models, including generating their own sets of focal variables[11,12]. Only one of our candidate mechanisms (diversity, but not processing capacity or emotional detachment) was directly measured, and none was independently manipulated, again calling for further research.

Finally, AI capabilities are advancing rapidly, and AI agents are likely to continue to improve at theory building alongside other scientific tasks. Future meta-scientific RCTs should randomly assign different stages of the scientific workflow to artificial intelligences vs. human researchers and examine the consequences for scientific productivity, quality, and intellectual diversity. Our materials are fully open source, and we strongly encourage a wave of future investigations building and improving on this exercise with more advanced artificial intelligences and further research domains.

# Extended Data

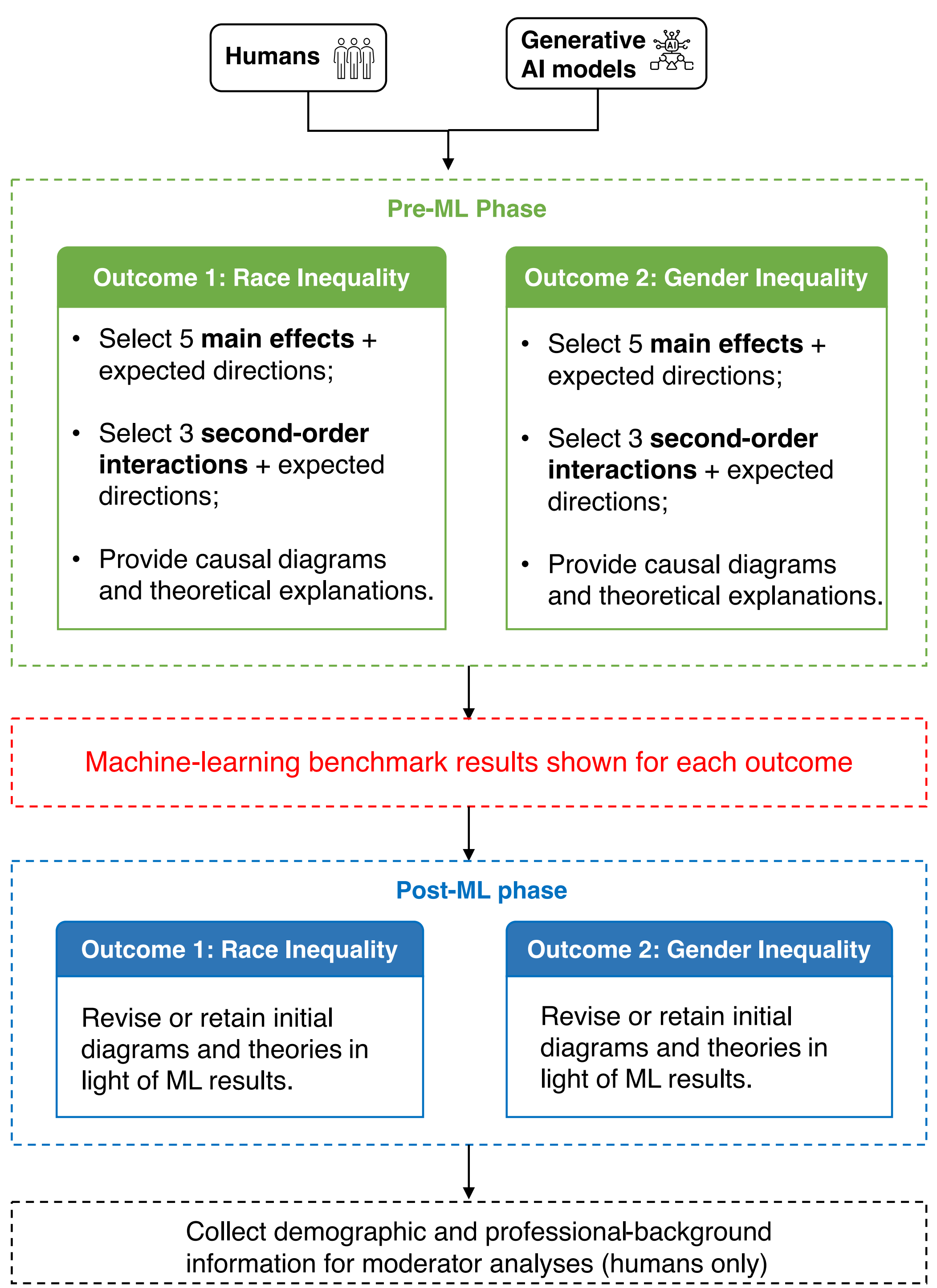

**Extended Data Fig. 1 | Survey procedure.** Contributors completed theory-generation and scientific prediction tasks before viewing the ML evidence and then revised or retained their theories. All contributors completed parallel racial- and gender-inequality tasks, with task order randomized across contributors. Human and GenAI contributors completed identical sets of questions.

**a. Mean prediction accuracy by group**

| Effect | Task | Group | N | Mean Accuracy | 95% CI | Aggregated Accuracy |
|---|---|---|---|---|---|---|
| Main Effects | Racial Inequality | Humans | 73 | 0.411 | [0.375, 0.449] | 0.584 |
| | | Senior Scientists | 13 | 0.415 | [0.354, 0.477] | 0.542 |
| | | PhD Students | 60 | 0.410 | [0.367, 0.453] | 0.592 |
| | | Topic Experts | 17 | 0.412 | [0.353, 0.471] | 0.532 |
| | | Non-Experts | 56 | 0.411 | [0.364, 0.457] | 0.599 |
| | | **GenAI** | 25 | **0.416** | [0.400, 0.440] | 0.475 |
| | Gender Inequality | Humans | 73 | 0.578 | [0.548, 0.608] | 0.731 |
| | | Senior Scientists | 13 | 0.585 | [0.554, 0.600] | 0.669 |
| | | PhD Students | 60 | 0.577 | [0.540, 0.610] | 0.744 |
| | | Topic Experts | 17 | 0.565 | [0.529, 0.600] | 0.683 |
| | | Non-Experts | 56 | 0.582 | [0.543, 0.618] | 0.744 |
| | | **GenAI** | 25 | **0.624** | [0.600, 0.648] | 0.693 |
| Second-Order Interactions | Racial Inequality | Humans | 73 | 0.128 | [0.059, 0.196] | 0.348 |
| | | Senior Scientists | 13 | 0.205 | [0.051, 0.385] | 0.381 |
| | | PhD Students | 60 | 0.111 | [0.039, 0.189] | 0.321 |
| | | Topic Experts | 17 | 0.196 | [0.059, 0.333] | 0.403 |
| | | Non-Experts | 56 | 0.107 | [0.030, 0.190] | 0.309 |
| | | **GenAI** | 25 | **0.400** | [0.320, 0.480] | 0.528 |
| | Gender Inequality | Humans | 73 | -0.005 | [−0.064, 0.055] | -0.009 |
| | | Senior Scientists | 13 | 0.051 | [−0.077, 0.179] | 0.078 |
| | | PhD Students | 60 | -0.017 | [−0.089, 0.050] | -0.036 |
| | | **Topic Experts** | 17 | **0.118** | [0.000, 0.235] | 0.179 |
| | | Non-Experts | 56 | -0.042 | [−0.113, 0.030] | -0.090 |
| | | GenAI | 25 | -0.013 | [−0.080, 0.053] | -0.017 |

**b. Two-sided tests of mean and aggregated prediction accuracy**

| Comparison | Effect | Task | $p_{mean}$ | $p_{agg}$ |
|---|---|---|---|---|
| Humans vs. GenAI | Main Effects | Racial Inequality | 0.820 | 0.023* |
| | | Gender Inequality | 0.026* | 0.206 |
| | Second-Order Interactions | Racial Inequality | <0.001*** | 0.174 |
| | | Gender Inequality | 0.856 | 1.000 |
| Topic Experts vs. GenAI | Main Effects | Racial Inequality | 0.902 | 0.180 |
| | | Gender Inequality | 0.016* | 0.766 |
| | Second-Order Interactions | Racial Inequality | 0.027* | 0.259 |
| | | Gender Inequality | 0.084 | 0.035* |
| Senior Scientists vs. GenAI | Main Effects | Racial Inequality | 0.987 | 0.132 |
| | | Gender Inequality | 0.062 | 0.521 |
| | Second-Order Interactions | Racial Inequality | 0.064 | 0.172 |
| | | Gender Inequality | 0.387 | 0.251 |
| PhD Students vs. Senior Scientists | Main Effects | Racial Inequality | 0.899 | 0.502 |
| | | Gender Inequality | 0.742 | 0.081 |
| | Second-Order Interactions | Racial Inequality | 0.347 | 0.792 |
| | | Gender Inequality | 0.366 | 0.491 |
| Topic Experts vs. Non-Experts | Main Effects | Racial Inequality | 0.979 | 0.318 |
| | | Gender Inequality | 0.523 | 0.095 |
| | Second-Order Interactions | Racial Inequality | 0.312 | 0.640 |
| | | Gender Inequality | 0.038* | 0.066 |

**Extended Data Table 1 | Prediction accuracy by contributor group for main effects and second-order interactions. a**, For each contributor group, sample size, mean accuracy, percentile-bootstrap 95% confidence interval, and aggregated accuracy. Within each effect-by-task block, the highest mean accuracy is shown in bold. **b**, Two-sided Welch's *t*-tests comparing different groups. *, **, and *** indicate $P < 0.05$, $P < 0.01$, and $P < 0.001$, respectively.

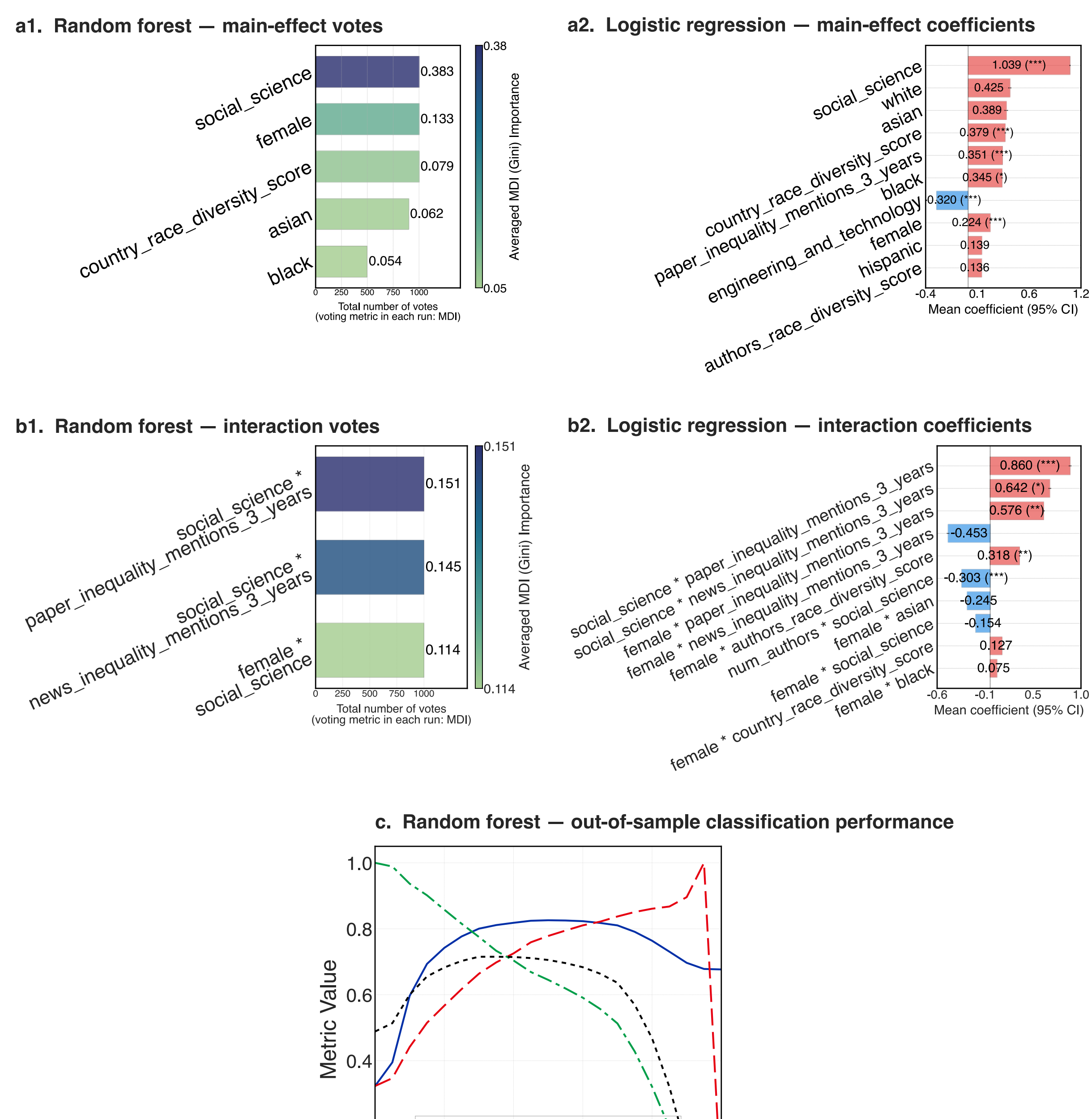

**Extended Data Fig. 2 | Machine-learning benchmark for racial-inequality discourse. a1, b1,** Features ranked by how often they appeared among the top five (**a1**) or top three (**b1**) predictors across random forest runs (bar length: vote count), using mean decrease in impurity (MDI). Numbers to the right of each bar, and the color scale, give the mean MDI of that feature across runs. **a2, b2,** Logistic-regression coefficients (mean across runs) with 95% confidence intervals; bar length shows the signed mean coefficient, and the number on each bar reports that mean (stars denote significance). Panels **a–b** aggregate 1,000 model runs: 100 random partitions of the full sample, each split into 10 subsamples, with a separate model fitted on each subsample. **c1**, Out-of-sample random-forest classification of racial-inequality papers: accuracy, precision, recall and F1 on a held-out test set (20% of the sample) as functions of the decision threshold; AUC is reported in the legend.

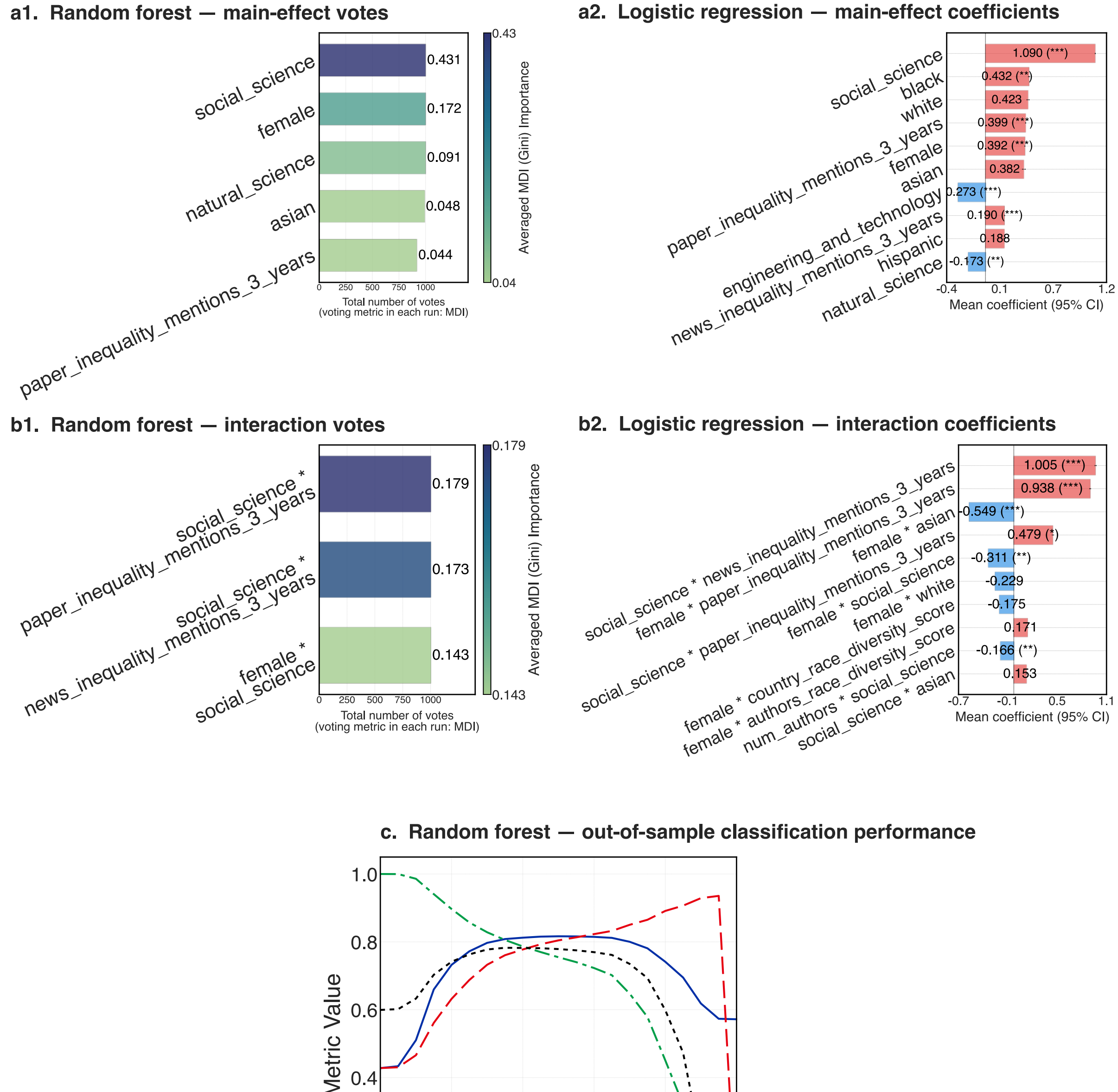

**Extended Data Fig. 3 | Machine-learning benchmark for gender-inequality discourse. a1, b1,** Features ranked by how often they appeared among the top five (**a1**) or top three (**b1**) predictors across random forest runs (bar length: vote count), using mean decrease in impurity (MDI). Numbers to the right of each bar, and the color scale, give the mean MDI of that feature across runs. **a2, b2,** Logistic-regression coefficients (mean across runs) with 95% confidence intervals; bar length shows the signed mean coefficient, and the number on each bar reports that mean (stars denote significance). Panels **a–b** aggregate 1,000 model runs: 100 random partitions of the full sample, each split into 10 subsamples, with a separate model fitted on each subsample. **c1**, Out-of-sample random-forest classification of gender-inequality papers: accuracy, precision, recall and F1 on a held-out test set (20% of the sample) as functions of the decision threshold; AUC is reported in the legend.

| Task | Prediction diversity | | | | Error diversity | | | |
|---|---|---|---|---|---|---|---|---|
| | Humans | GenAI | $P$ | $\hat{\gamma}_P$ | Humans | GenAI | $P$ | $\hat{\gamma}_E$ |
| *Main effects* | | | | | | | | |
| Racial inequality | 0.511 [0.450, 0.560] | 0.243 [0.185, 0.283] | <0.001*** | 0.310*** [0.296, 0.324] | 0.433 [0.374, 0.481] | 0.208 [0.155, 0.242] | <0.001*** | 0.356*** [0.342, 0.370] |
| Gender inequality | 0.380 [0.324, 0.426] | 0.197 [0.121, 0.250] | <0.001*** | 0.325*** [0.316, 0.335] | 0.436 [0.384, 0.477] | 0.267 [0.163, 0.347] | <0.001*** | 0.259*** [0.251, 0.267] |
| *Interactions* | | | | | | | | |
| Racial inequality | 0.877 [0.821, 0.908] | 0.444 [0.332, 0.514] | <0.001*** | 0.387*** [0.355, 0.418] | 0.511 [0.466, 0.542] | 0.340 [0.262, 0.391] | <0.001*** | 0.529*** [0.499, 0.559] |
| Gender inequality | 0.765 [0.684, 0.820] | 0.429 [0.337, 0.481] | <0.001*** | 0.017** [0.007, 0.028] | 0.378 [0.331, 0.411] | 0.203 [0.163, 0.219] | <0.001*** | 0.076*** [0.057, 0.094] |

**Extended Data Table 2 | Prediction heterogeneity and its association with aggregation gain.** The Humans and GenAI columns report mean within-group prediction diversity or error diversity, as indicated by the column heading; higher values indicate greater diversity. Values in brackets for these diversity measures are 95% confidence intervals from contributor-level bootstrap resampling (2,000 resamples). $P$ values compare Humans and GenAI in terms of prediction diversity and error diversity, using two-sided contributor-level permutation tests with 5,000 permutations. The coefficients $\hat{\gamma}_P$ and $\hat{\gamma}_E$ estimate the associations of prediction diversity and error diversity, respectively, with aggregation gains. Separate OLS models were estimated for each task and effect type:

$$\Delta_c = \beta_0 + \hat{\gamma}_P D_c^P + \beta_1 \text{Human}_c + \beta_2 \bar{a}_c + \beta_3 k_c + \varepsilon_c,$$

and

$$\Delta_c = \beta_0 + \hat{\gamma}_E D_c^E + \beta_1 \text{Human}_c + \beta_2 \bar{a}_c + \beta_3 k_c + \varepsilon_c.$$

Here, $c$ indexes a resampled crowd; $\Delta_c$ is aggregation gain; $D_c^P$ and $D_c^E$ are prediction diversity and error diversity for crowd $c$, respectively; $\text{Human}_c$ equals 1 for human crowds and 0 for GenAI crowds; $\bar{a}_c$ is mean individual prediction accuracy within the crowd; and $k_c$ is crowd size. Models were estimated using equal-sized crowds of $k = 2, \ldots, 25$, with 200 draws at each crowd size. Values in brackets for $\hat{\gamma}_P$ and $\hat{\gamma}_E$ are 95% confidence intervals calculated using the corresponding HC1-robust standard errors. Significance stars are based on the associated HC1-robust $P$ values.

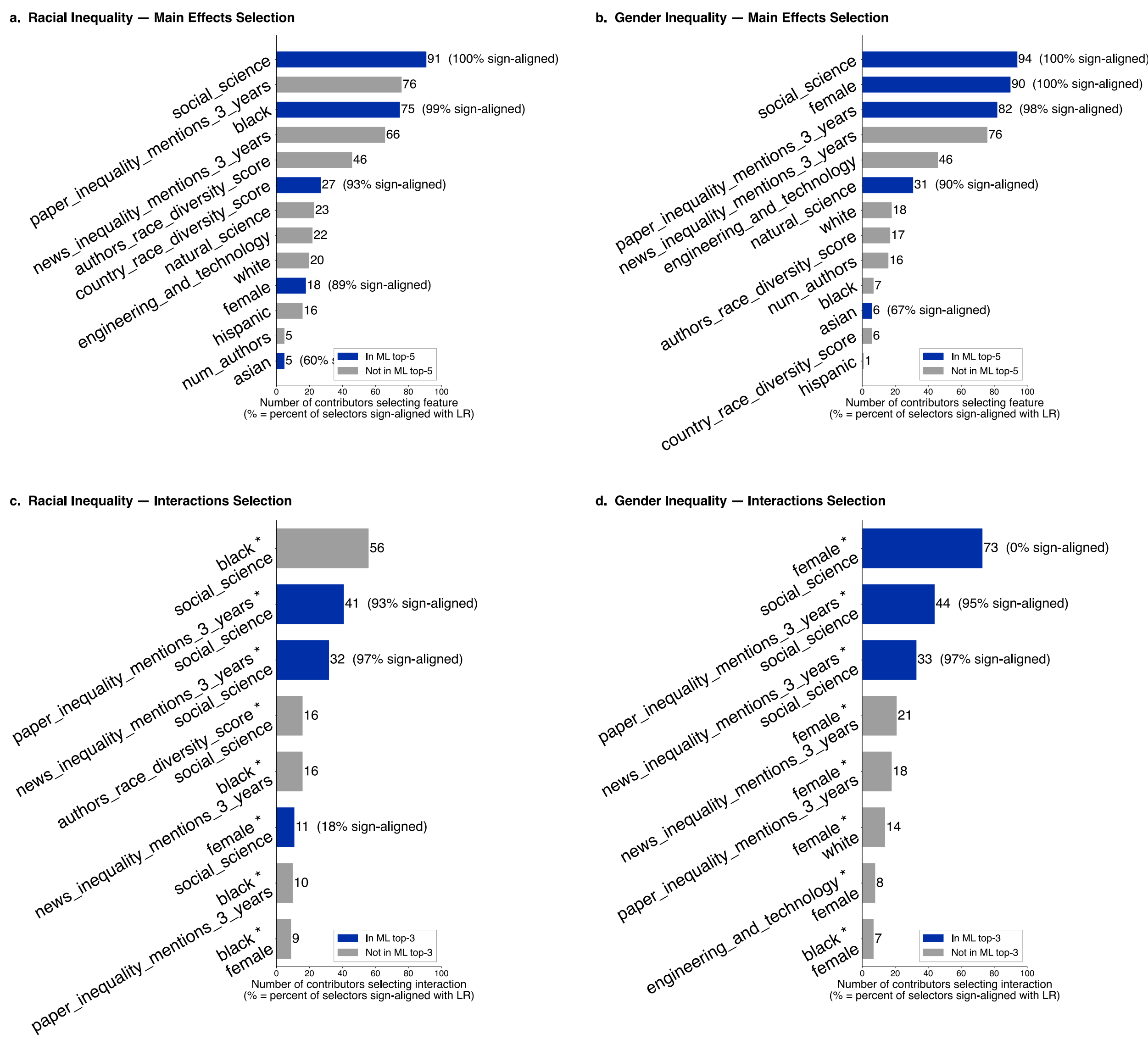

**Extended Data Fig. 4 | Theorists' selection of predictors and alignment with benchmark directions. a,b,** Main-effect predictions for racial- and gender-inequality discourse, respectively. **c,d,** Second-order interaction predictions for racial- and gender-inequality discourse, respectively. Bars show the number of contributors who selected each feature or interaction, ordered by selection frequency. Blue bars denote features ranked by the ML benchmark among the five strongest predictors among all the main effects, or interactions ranked among the three strongest predictors among all the interaction effects; grey bars denote other selected features or interactions. Numbers at the ends of bars indicate the number of theorists selecting each item. For benchmark-selected predictors shown in blue, percentages in parentheses indicate the proportion of selectors whose predicted direction matched the corresponding logistic-regression coefficient. Full details of features are provided in the Materials and Methods section.

| Task | Forecast type | ***Female*** **($\beta_1$)** | ***Expertise*** **($\beta_2$)** | ***Seniority*** **($\beta_3$)** |
|---|---|---|---|---|
| Racial inequality | Main effects | 0.047<br>$[-0.041,\ 0.136]$ | -0.004<br>$[-0.083,\ 0.075]$ | 0.004<br>$[-0.078,\ 0.086]$ |
| | Interactions | -0.065<br>$[-0.209,\ 0.079]$ | 0.058<br>$[-0.185,\ 0.301]$ | 0.057<br>$[-0.215,\ 0.328]$ |
| Gender inequality | Main effects | 0.019<br>$[-0.039,\ 0.076]$ | -0.041<br>$[-0.107,\ 0.026]$ | 0.037<br>$[-0.021,\ 0.095]$ |
| | Interactions | 0.088<br>$[-0.041,\ 0.218]$ | 0.217*<br>$[-0.002,\ 0.436]$ | -0.101<br>$[-0.351,\ 0.148]$ |

**Extended Data Table 3 | Associations between contributor characteristics and predictive accuracy.** Analyses include human scientists only (N = 73). The dependent variable is predictive accuracy. Within each task × effect cell, estimates are from the multivariate OLS regression: $\text{Prediction Accuracy}_i = \beta_0 + \beta_1 * \text{Female}_i + \beta_2 * \text{Expertise}_i + \beta_3 * \text{Seniority}_i + \varepsilon_i$. $\text{Female}_i$ is coded 1 for female contributors and 0 for other contributors. $\text{Expertise}_i$ is coded 1 for contributors with at least one peer-reviewed publication on racial or gender inequality and 0 otherwise. $\text{Seniority}_i$ is coded 1 for senior scientists and 0 for doctoral students. Reported cells are the coefficients $\beta_1$, $\beta_2$, and $\beta_3$ with HC1 95% CIs. Significance stars are based on one-sided HC1 $t$-tests of the directional alternative that the indicated characteristic is associated with greater predictive accuracy. * indicates $P < 0.05$.

**a. Theory complexity by group**

| Phase | Task | Metric | Group means | | | | | | Two-sided $P$ |
|---|---|---|---|---|---|---|---|---|---|
| | | | Humans | Senior | PhD | Experts | Non-Experts | GenAI | Humans vs GenAI |
| Pre-ML | Racial inequality | No. of paths | 5.18 | 5.17 | 5.19 | 5.20 | 5.18 | 6.64 | $0.014^{*}$ |
| | | Maximum path length | 2.14 | 2.00 | 2.17 | 1.87 | 2.21 | 2.64 | $0.004^{**}$ |
| | | No. of latent variables | 2.15 | 2.00 | 2.19 | 1.67 | 2.29 | 5.16 | $< 0.001^{***}$ |
| | | Theory word count | 125.56 | 129.50 | 124.75 | 133.07 | 123.54 | 273.44 | $< 0.001^{***}$ |
| | Gender inequality | No. of paths | 5.24 | 5.23 | 5.24 | 5.12 | 5.27 | 6.80 | $0.027^{*}$ |
| | | Maximum path length | 2.03 | 2.08 | 2.02 | 2.00 | 2.04 | 2.60 | $0.001^{**}$ |
| | | No. of latent variables | 1.94 | 1.85 | 1.97 | 1.62 | 2.04 | 5.52 | $< 0.001^{***}$ |
| | | Theory word count | 119.82 | 124.12 | 118.87 | 116.78 | 120.69 | 300.50 | $< 0.001^{***}$ |
| Post-ML | Racial inequality | No. of paths | 5.09↓ | 4.25↓ | 5.26↑ | 4.40↓ | 5.28↑ | 8.28↑ | $0.002^{**}$ |
| | | Maximum path length | 2.00↓ | 1.75↓ | 2.05↓ | 1.80↓ | 2.06↓ | 2.68↑ | $0.001^{**}$ |
| | | No. of latent variables | 2.03↓ | 2.08↑ | 2.02↓ | 1.87↑ | 2.07↓ | 6.72↑ | $< 0.001^{***}$ |
| | | Theory word count | 131.88↑ | 152.50↑ | 127.54↑ | 146.77↑ | 127.74↑ | 320.44↑ | $< 0.001^{***}$ |
| | Gender inequality | No. of paths | 5.11↓ | 4.77↓ | 5.19↓ | 4.69↓ | 5.23↓ | 8.56↑ | $0.010^{*}$ |
| | | Maximum path length | 1.99↓ | 1.85↓ | 2.02 | 1.88↓ | 2.02↓ | 2.60 | $0.001^{**}$ |
| | | No. of latent variables | 1.94 | 1.54↓ | 2.03↑ | 1.38↓ | 2.11↑ | 7.28↑ | $< 0.001^{***}$ |
| | | Theory word count | 125.49↑ | 124.04↓ | 125.81↑ | 117.88↑ | 127.66↑ | 357.80↑ | $< 0.001^{***}$ |

**b. Association between complexity and prediction accuracy**

| Complexity metric ($C$) | $\beta_1$ ($C$) | $\beta_2$ (GenAI) | $\beta_3$ ($C\times$GenAI) | Human slope | GenAI slope |
|---|---|---|---|---|---|
| Number of paths | 0.009<br>[−0.021, 0.039] | 0.023<br>[−0.010, 0.056] | −0.007<br>[−0.042, 0.028] | 0.009<br>[−0.021, 0.039] | 0.002<br>[−0.017, 0.021] |
| Maximum path length | 0.020<br>[−0.007, 0.048] | 0.032<br>[−0.007, 0.071] | $-0.040^{*}$<br>[−0.075, −0.005] | 0.020<br>[−0.007, 0.048] | −0.020<br>[−0.041, 0.002] |
| Number of latent variables | 0.016<br>[−0.016, 0.048] | 0.031<br>[−0.011, 0.072] | −0.028<br>[−0.065, 0.009] | 0.016<br>[−0.016, 0.048] | −0.012<br>[−0.030, 0.007] |
| Theory word count | −0.013<br>[−0.054, 0.029] | $0.086^{***}$<br>[0.043, 0.129] | 0.013<br>[−0.034, 0.060] | −0.013<br>[−0.054, 0.029] | −0.000<br>[−0.022, 0.022] |

**c. Association between complexity and LLM-assessed theory quality ratings**

| Complexity metric ($C$) | $\beta_1$ ($C$) | $\beta_2$ (GenAI) | $\beta_3$ ($C\times$GenAI) | Human slope | GenAI slope |
|---|---|---|---|---|---|
| Number of paths | $0.531^{**}$<br>[0.213, 0.849] | $2.535^{***}$<br>[2.176, 2.894] | −0.301<br>[−0.649, 0.048] | $0.531^{**}$<br>[0.213, 0.849] | $0.231^{**}$<br>[0.090, 0.371] |
| Maximum path length | −0.006<br>[−0.285, 0.274] | $2.661^{***}$<br>[2.276, 3.047] | 0.222<br>[−0.108, 0.552] | −0.006<br>[−0.285, 0.274] | $0.216^{*}$<br>[0.042, 0.391] |
| Number of latent variables | 0.336<br>[−0.081, 0.753] | $2.522^{***}$<br>[2.132, 2.912] | −0.169<br>[−0.614, 0.275] | 0.336<br>[−0.081, 0.753] | $0.166^{*}$<br>[0.021, 0.312] |
| Theory word count | $2.154^{***}$<br>[1.622, 2.687] | $2.219^{***}$<br>[1.900, 2.538] | $-1.935^{***}$<br>[−2.472, −1.397] | $2.154^{***}$<br>[1.622, 2.687] | $0.220^{***}$<br>[0.135, 0.304] |

**d. Performance–complexity ratios (higher = more efficient)**

| Comparison | Complexity metric ($C$) | Accuracy / $C$ | | | Quality rating / $C$ | | |
|---|---|---|---|---|---|---|---|
| | | Group 1 | Group 2 | Two-sided $P$ | Group 1 | Group 2 | Two-sided $P$ |
| Humans (Group 1) vs. GenAI (Group 2) | Number of paths | 0.115 | 0.086 | $< 0.001^{***}$ | 1.261 | 1.255 | 0.914 |
| | Maximum path length | 0.270 | 0.219 | $0.006^{**}$ | 3.122 | 3.414 | 0.052 |
| | Number of latent variables | 0.187 | 0.102 | $< 0.001^{***}$ | 2.090 | 1.800 | 0.172 |
| | Theory word count | 0.003 | 0.001 | $< 0.001^{***}$ | 0.051 | 0.032 | $< 0.001^{***}$ |
| Senior (Group 1) vs. PhD (Group 2) | Number of paths | 0.111 | 0.116 | 0.663 | 1.483 | 1.214 | 0.056 |
| | Maximum path length | 0.268 | 0.271 | 0.909 | 3.620 | 3.016 | $0.022^{*}$ |
| | Number of latent variables | 0.180 | 0.189 | 0.729 | 2.243 | 2.059 | 0.394 |
| | Theory word count | 0.004 | 0.003 | 0.432 | 0.054 | 0.050 | 0.127 |
| Experts (Group 1) vs. Non-Experts (Group 2) | Number of paths | 0.112 | 0.116 | 0.764 | 1.454 | 1.207 | $0.040^{*}$ |
| | Maximum path length | 0.273 | 0.270 | 0.884 | 3.587 | 2.993 | $0.010^{**}$ |
| | Number of latent variables | 0.194 | 0.186 | 0.764 | 2.260 | 2.049 | 0.306 |
| | Theory word count | 0.003 | 0.003 | 0.870 | 0.054 | 0.050 | 0.065 |

**Extended Data Table 4 | Theory complexity, prediction accuracy, and theory quality ratings. a,** Between-group comparisons of theory complexity across phases. In the post-ML panel, ↑/↓ mark descriptively whether each group mean increased or decreased relative to the same group's pre-ML mean for that metric. b, Interaction models relating pre-ML theory complexity ($\boldsymbol{C}$) to prediction accuracy: $\text{Prediction Accuracy} = \beta_0 + \beta_1 C + \beta_2 \text{GenAI} + \beta_3 (C \times \text{GenAI}) + \delta_{\text{task}\times\text{effect}} + \varepsilon$. c, Interaction models relating phase-matched theory complexity ($C$) to LLM-assessed theory-quality ratings:
$\text{Theory Quality Rating} = \beta_0 + \beta_1 C + \beta_2 \text{GenAI} + \beta_3 (C \times \text{GenAI}) + \delta_{\text{task}\times\text{effect}\times\text{phase}} + \varepsilon$. In **b** and **c**, $\beta_1$ is the slope for human contributors and $\beta_1 + \beta_3$ is the slope for GenAI contributors. Values in brackets are 95% confidence intervals. Standard errors and $P$ values for all regression coefficients use CR1 cluster-robust inference. $\delta_{\text{task}\times\text{effect}}$ and $\delta_{\text{task}\times\text{effect}\times\text{phase}}$ are fixed effects for the cells included in each model. For causal-diagram models, those cells are main-effects only (diagrams were collected for main effects only); for theory word-count models, cells include both main-effects and second-order-interaction outcomes.
**d,** Performance–complexity ratios. Cells report the mean ratio of prediction accuracy (left) or LLM-assessed theory-quality ratings (right) to the indicated theory-complexity metric $C$. For these ratios, pre-ML theory complexity is used for prediction accuracy; phase-matched theory complexity is used for quality ratings. Higher values indicate higher performance per unit complexity. $P$ values are from two-sided Welch $t$-tests comparing Group 1 with Group 2. *, **, and *** indicate $P < 0.05$, $P < 0.01$, and $P < 0.001$, respectively.

**a. Core fraction of theoretical explanations before and after ML exposure**

| Group | Topic | Effect | Core fraction (%) | | No. of cores | |
|---|---|---|---|---|---|---|
| | | | Pre-ML | Post-ML | Pre-ML | Post-ML |
| Humans | Race | Main effects | 17.8 [10.7, 28.1] | 31.5↑ [22.0, 42.9] | 1 | 1 |
| | | Interactions | 13.7 [7.6, 23.4] | 23.3↑ [15.1, 34.2] | 1 | 1 |
| | Gender | Main effects | 9.6 [4.7, 18.5] | 30.1↑ [20.8, 41.4] | 1 | 1 |
| | | Interactions | 12.3 [6.6, 21.8] | 21.9↑ [14.0, 32.7] | 1 | 1 |
| GenAI | Race | Main effects | 100.0 [86.7, 100.0] | 100.0 [86.7, 100.0] | 1 | 1 |
| | | Interactions | 92.0 [75.0, 97.8] | 100.0↑ [86.7, 100.0] | 1 | 1 |
| | Gender | Main effects | 92.0 [75.0, 97.8] | 100.0↑ [86.7, 100.0] | 1 | 1 |
| | | Interactions | 92.0 [75.0, 97.8] | 100.0↑ [86.7, 100.0] | 1 | 1 |

**b. Sensitivity of HDBSCAN core fraction to minimum cluster size (2–10)**

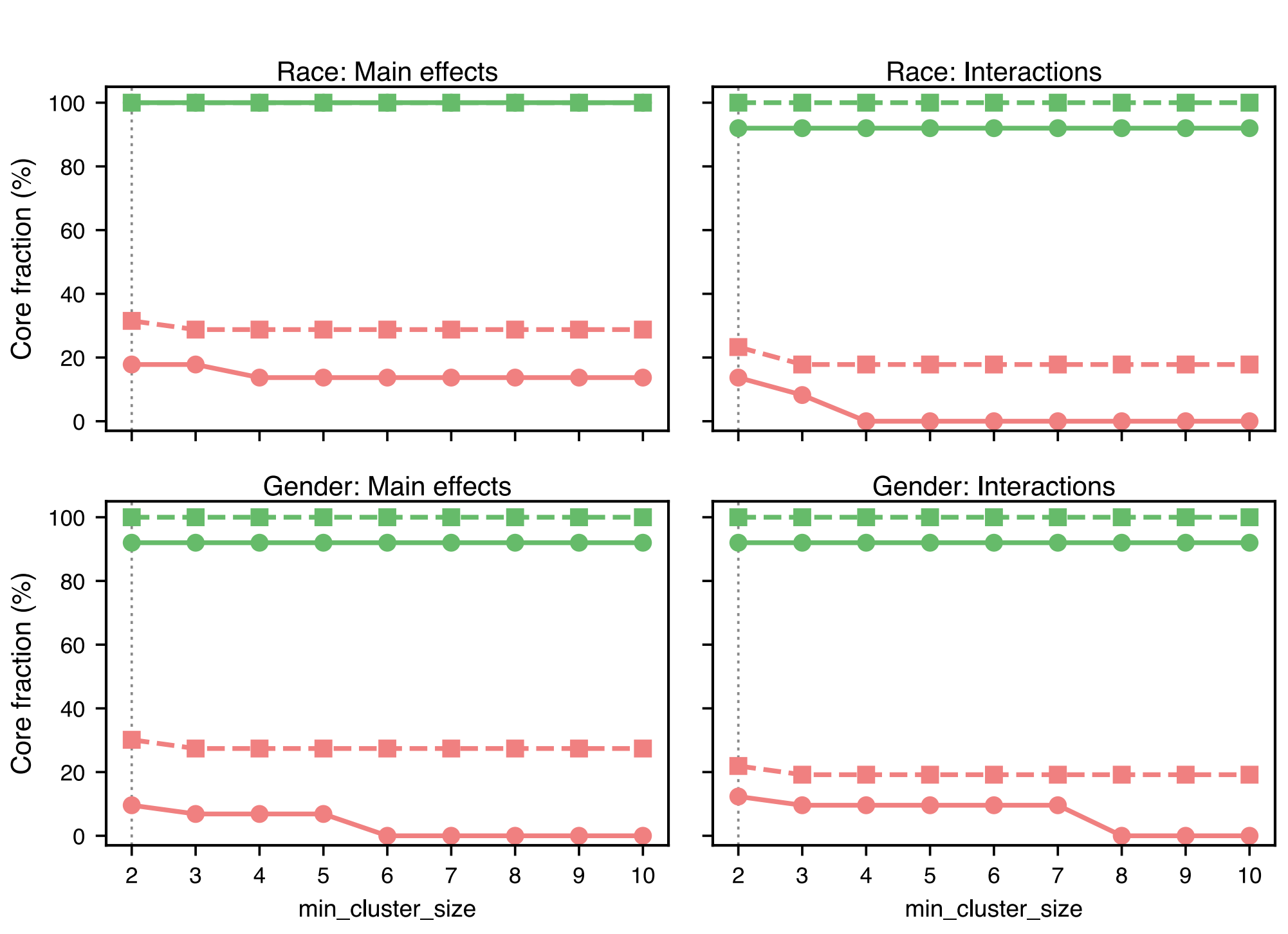

**Extended Data Fig. 5** | Core–tail structure of theoretical explanations and robustness to HDBSCAN parameters. **a,** Core fractions were estimated within each group, topic, and effect condition using within-period HDBSCAN. "No. of cores" is the number of HDBSCAN clusters among core explanations. ↑/↓ in the Post-ML column descriptively indicate the direction of change relative to Pre-ML. Values in brackets are 95% confidence intervals. **b,** Sensitivity of estimated core fractions to the HDBSCAN minimum cluster size, varied from two to 10. Solid lines denote pre-ML estimates and dashed lines post-ML estimates. The vertical dotted line indicates the value used in the primary analysis (min_cluster_size = 2).

**a. Do human and LLM evaluators agree on theory quality?**

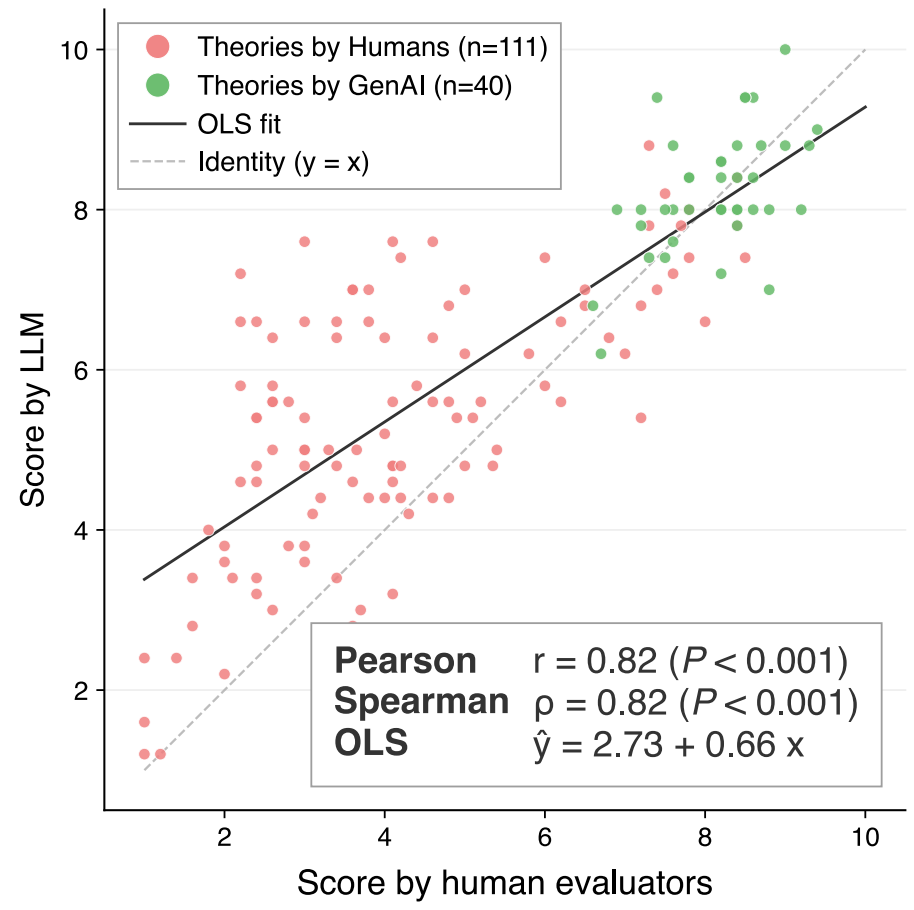

**b. Independent evaluations of theory quality (left: LLM evaluator; right: human evaluators)**

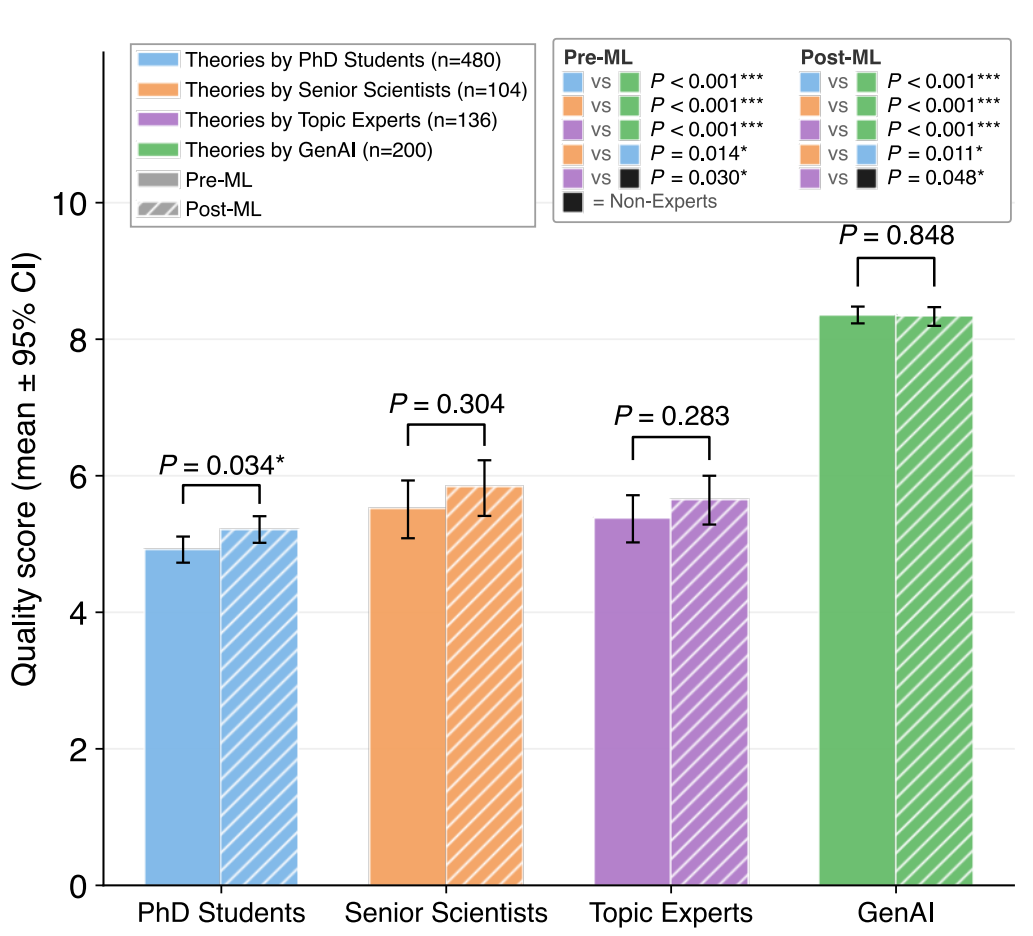

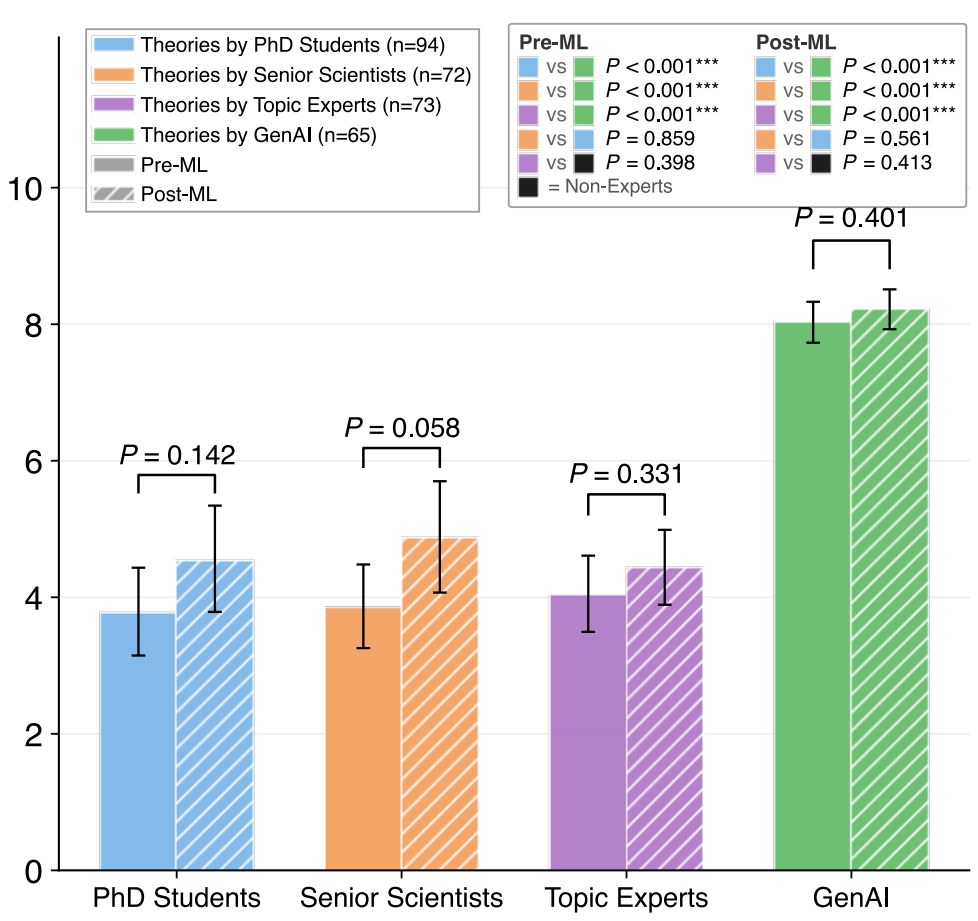

**c. Score distributions of theory quality (left: LLM evaluator; right: human evaluators)**

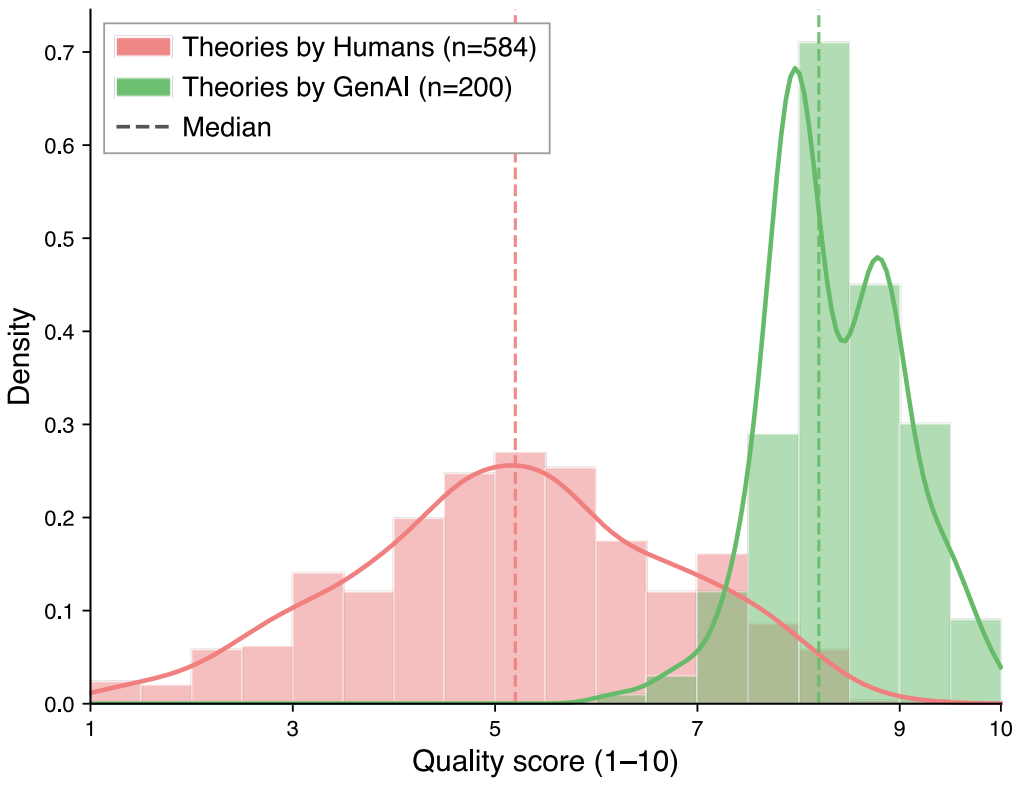

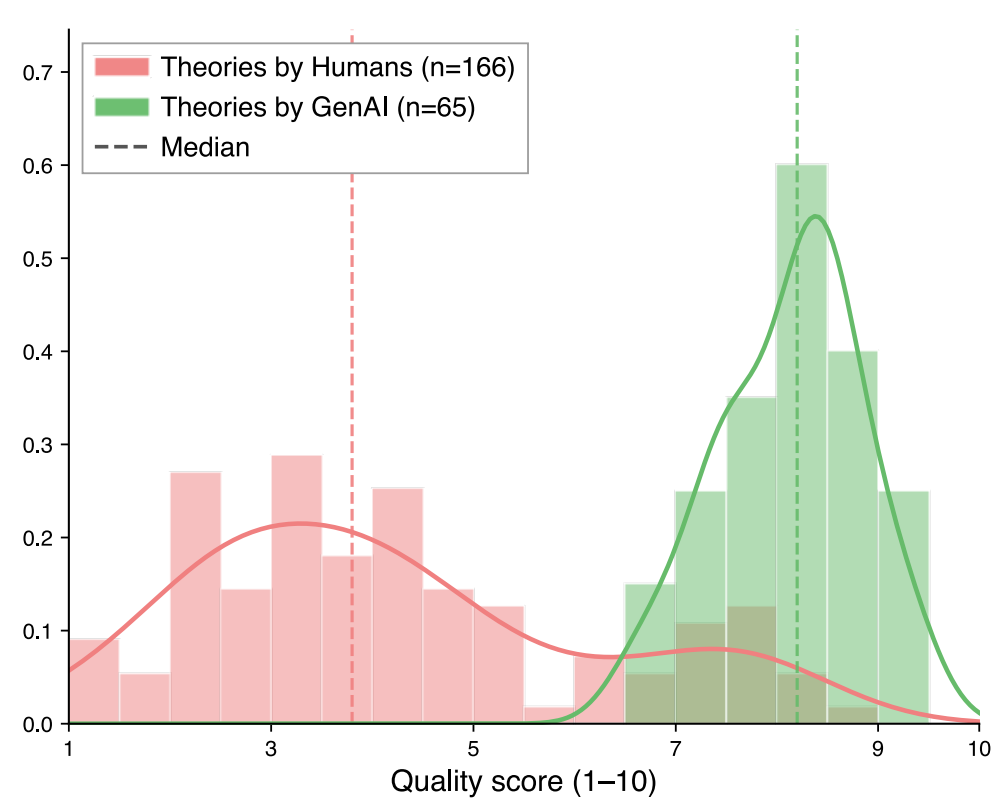

**Extended Data Fig. 6 | GenAI theories receive higher quality ratings across human and LLM evaluators**. **a,** Agreement between quality ratings from human independent raters and the LLM evaluator. Each point represents one theory generated by a human contributor or a GenAI system. Pearson $r$ and Spearman $\rho$ are shown with two-sided $P$ values. **b,** Mean quality ratings before (solid) and after (hatched) exposure to the ML results; error bars show 95% bootstrap confidence intervals from 5,000 resamples. Brackets compare pre- versus post-ML theory quality ratings within contributor groups; the upper-right inset shows between-group differences in theory quality ratings within each phase. All comparisons use two-sided Welch's $t$-tests. **c,** Distributions of quality scores. Histograms are overlaid with kernel density estimates; dashed lines indicate medians.

# Methods

## M.1 Study design

**Pre-registration.** The preregistered analysis plan for this investigation is available at https://osf.io/buah6/overview?view_only=98ca17155c4d4463acb6f7c043ba54b1. We compared the capabilities of human researchers and generative AI systems on core components of scientific theorizing: anticipating empirical patterns, proposing explanations for those patterns, and revising explanations in response to new evidence. Our set of pre-registered statistical analyses afforded tests of competing theories of AI strengths and weaknesses as outlined in the introduction. However, our approach was fundamentally abductive and focused on inference to the best currently available explanation[30], as opposed to carrying out confirmatory tests of directional predictions on the part of the core team. Deviations from and additions to the originally planned analyses are summarized in the Supplementary Information.

**Contributors.** The human contributors comprised 13 senior scientists (professors) and 60 PhD students. Senior scientists were recruited through the professional networks of the research team. Eligibility for senior faculty required contributors to be tenured or tenure-track faculty members at leading academic institutions and to have research expertise broadly related to prejudice, stereotyping, discrimination, inequality, or social judgments. PhD students were recruited through doctoral-program and laboratory directors at leading institutions, with no restrictions on topic specialization. In recognition of their substantive contributions to theory development, human contributors were offered authorship on the study. The GenAI contributors comprised 25 widely used generative AI models available at the time of the study. Models were accessed through their standard conversational interfaces rather than APIs, allowing information to be presented sequentially while preserving prior survey context. Each model completed the survey in a separate conversation, constituting an independent run. Demographic details regarding the human contributors and the full list of GenAI models are provided in the Supplementary Information.

**Survey.** All contributors completed two parallel tasks, one for racial inequality and one for gender inequality, with task order randomized across respondents. For each outcome, contributors selected the five main effects and three second-order interactions they expected to be most predictive and specified the expected direction of each association. They then provided a written theoretical explanation (for main effects and second-order interactions) and a corresponding causal diagram (for main effects). They were then shown the corresponding ML evidence for each outcome and could revise their written explanations and causal diagrams or retain their initial theories if they judged that revision was not warranted. The survey procedure is presented in Extended Data Fig. 1. Human participants and GenAI systems received identical survey instructions. The complete survey is provided in the Supplementary Information.

## M.2 Discovering empirical patterns using machine-learning algorithms

**Dataset.** We constructed the empirical corpus from a Semantic Scholar database maintained in SOLR by Professor Philip M. Parker's data team at INSEAD. The database contains approximately 180 million research papers with metadata including titles, abstracts, authors, publication venues, and years. Using expert-generated keyword dictionaries, we identified papers published between 2005 and 2021 that contained at least one racial- or gender-inequality keyword in the title or abstract. We also sampled approximately 50,000 papers containing no keywords from either dictionary as a shared negative set. This yielded two classification datasets: a racial inequality dataset of approximately 80,000 papers, comprising racial inequality papers and the shared negative set, and a gender-inequality dataset of approximately 95,000 papers, comprising gender-inequality papers and the same negative set. The complete keyword dictionaries are provided in the Supplementary Information.

**Features.** For each paper, we extracted 13 preregistered features capturing the disciplinary domain, author demographics, author diversity, and time-lagged measures of the prevalence of inequality discourses. Full definitions and construction procedures are provided in the Supplementary Information.

**Machine-learning benchmarks.** We used supervised machine-learning models to identify empirical predictors of racial- and gender-inequality discourse at the paper level. Separate models were estimated for each outcome. Main-effect models used the 13 preregistered predictors, whereas interaction models used all 78 pairwise interactions among these predictors without including the corresponding main effects. Random-forest[70] classifiers were used to rank predictors, with random-forest hyperparameters selected by cross-validation. To obtain stable rankings, we repeatedly reshuffled and subsampled the data, yielding 1,000 fitted models for each outcome and model specification. Predictors were ranked according to how consistently they appeared among the most important features across runs; the top five main effects and top three interactions were retained as the ML benchmarks. We then used logistic regression[71] to estimate the direction of association for the selected main effects and interactions, again using repeated subsampling. The machine-learning benchmark results are reported in Extended Data Fig. 2 and Extended Data Fig. 3. Full model specifications, finetuning procedures and robustness checks are provided in the Supplementary Information.

## M.3 Analyses of scientific predictions

Unless otherwise noted, contributor-level measures were computed separately within each task-by-effect condition.

**Accuracy of scientific predictions.** We measured prediction accuracy by comparing each contributor's predictions with the corresponding machine-learning benchmark using cosine similarity. For main-effect predictions, each contributor's response was represented as a 13-dimensional signed prediction vector,

$$\boldsymbol{P}_i = (P_{i,1}, \dots, P_{i,13}),$$

where $i$ indexes contributors and each dimension corresponds to one candidate feature. An entry was coded as +1 if the contributor selected the feature and predicted a positive association, -1 if the contributor selected the feature and predicted a negative association, and 0 if the contributor did not select the feature. Because contributors selected five features, each main-effect prediction vector contained five nonzero entries.

The corresponding machine-learning benchmark vector $\boldsymbol{b}$ was encoded in the same way, in which the five features selected by the random-forest analysis were assigned values of $+1$ or $-1$ according to the direction estimated by logistic regression; all remaining entries were coded as $0$.

Contributor $i$'s prediction accuracy was defined as the cosine similarity between the contributor's signed prediction vector and the corresponding machine-learning benchmark vector:

$$a_i = \frac{\boldsymbol{P}_i \cdot \boldsymbol{b}}{\| \boldsymbol{P}_i \|_2 \| \boldsymbol{b} \|_2}.$$

Scores range from -1 to 1, with higher values indicating greater agreement with the benchmark in both feature selection and predicted direction.

Interaction predictions were scored analogously using 78-dimensional vectors representing all pairwise combinations of the 13 features, with three nonzero entries corresponding to the selected interactions.

Let $G_g$ denote the set of contributors in group $g$ (human or GenAI), and $n_g = |G_g|$. Mean individual prediction accuracy within the same group was

$$\bar{a}_g = \frac{1}{n_g} \sum_{i \in G_g} a_i.$$

Mean individual prediction accuracy was compared between human and GenAI contributors using two-sided Welch's *t*-tests. Group means are shown with 95% bootstrap confidence intervals based on 5,000 resamples.

**Random benchmark.** Consider selecting five main effects or three interactions uniformly at random, assigning each a random direction (+1 or −1), and calculating cosine similarity with the ML benchmark. The theoretical expectation of the resulting cosine similarity is then zero, which we use as the random benchmark.

**Aggregation.** We constructed the aggregated predictions for group $g$ by averaging the signed prediction vectors of its members,

$$\bar{\boldsymbol{P}}_g = \frac{1}{n_g} \sum_{i \in G_g} \boldsymbol{P}_i.$$

The accuracy of the aggregated predictions was then defined as

$$A_g = \frac{\bar{\boldsymbol{P}}_g \cdot \boldsymbol{b}}{\| \bar{\boldsymbol{P}}_g \|_2 \| \boldsymbol{b} \|_2}.$$

We defined aggregation gain as the gap between accuracy of the aggregated predictions and mean individual accuracy within the same contributor group:

$$\Delta_g = A_g - \bar{a}_g.$$

To characterize aggregation accuracy and aggregation gain across crowd sizes, we repeatedly sampled contributors without replacement within each group, with crowd size *k* ranging from two to that group's pool size. For each *k*, we sampled 1,000 crowds and summarized aggregated accuracy and aggregation gain by their means and 2.5th–97.5th percentiles.

**Heterogeneity in scientific predictions.** We characterized within-group prediction heterogeneity using two complementary measures: prediction diversity and error diversity. For contributor group $g$, prediction similarity was defined as the mean pairwise cosine similarity between contributors' signed prediction vectors:

$$S_g^P = \frac{1}{N_{\text{pairs},g}} \sum_{\substack{i<j \\ i,j\in G_g}} \frac{\boldsymbol{P}_i \cdot \boldsymbol{P}_j}{\| \boldsymbol{P}_i \|_2 \| \boldsymbol{P}_j \|_2},$$

where $N_{\text{pairs},g} = \binom{n_g}{2}$ is the number of unique contributor pairs in group $g$. Prediction diversity was defined as

$$D_g^P = 1 - S_g^P.$$

To characterize heterogeneity in prediction errors, we defined contributor $i$'s error vector relative to the corresponding machine-learning benchmark as

$$\boldsymbol{e}_i = \boldsymbol{P}_i - \boldsymbol{b}.$$

Error similarity within group $g$ was defined as the mean pairwise cosine similarity between contributors' error vectors:

$$S_g^E = \frac{1}{N_{\text{pairs},g}} \sum_{\substack{i<j \\ i,j\in G_g}} \frac{\boldsymbol{e}_i \cdot \boldsymbol{e}_j}{\| \boldsymbol{e}_i \|_2 \| \boldsymbol{e}_j \|_2},$$

and error diversity as

$$D_g^E = 1 - S_g^E.$$

Higher values of $D_g^P$ indicate greater heterogeneity in predictions, whereas higher values of $D_g^E$ indicate greater heterogeneity in prediction errors.

### M.4 Analyses of theories

**Theory complexity.** We quantified the structural theory complexity using four measures: the number of causal paths, the maximum path length, the number of latent variables, based on the provided causal diagram; and the theory word count, based on the provided textual explanation. The first three were extracted from the submitted diagrams using GPT-5.5 with high reasoning effort, with a small subsample independently coded by a human coder showing near perfect agreement with the model-based measures. The theory word count was the number of whitespace-separated tokens in the corresponding theoretical explanation. Between-group differences in complexity were evaluated within each phase using two-sided Welch's *t*-tests. The extraction prompts and human validation of diagram complexity are provided in the Supplementary Information.

**Semantic representation and diversity.** We represented each theoretical explanation as a 3,072-dimensional embedding using OpenAI's pretrained *text-embedding-3-large* model. For visualization, embeddings were projected into two dimensions using PCA; all quantitative analyses used the original 3,072-dimensional embedding space. Semantic diversity was measured as (i) the cosine distance (defined as 1 – cosine similarity) of each explanation from its group centroid and (ii) the mean pairwise cosine distance among explanations within each group, with larger values indicating greater semantic dispersion. Analyses were pooled across task-by-effect conditions. For centroid distance, between-group contrasts among human subgroups (PhD students vs. senior scientists) used two-sided Welch's $t$-tests; the Human–GenAI contrast used a one-sided Welch's $t$-test ($H_1$: Human > GenAI); and within-group pre- to post-ML changes used paired one-sided $t$-tests ($H_1$: post < pre). For mean pairwise cosine distance, between-group contrasts among human subgroups used two-sided permutation tests (5,000 permutations); the Human–GenAI contrast used one-sided permutation tests (5,000 permutations; $H_1$: Human > GenAI); and within-group pre- to post-ML changes used paired one-sided permutation tests (5,000 permutations; $H_1$: post < pre). Group means were summarized with bootstrap 95% confidence intervals.

**Core–tail structure.** We further characterized the distributional structure of theoretical explanations using HDBSCAN clustering[72] with cosine distance as the metric, fitted separately within each contributor group, task and phase. Clustering was performed on L2-normalized 3,072-dimensional embeddings using the Python *hdbscan* package. Explanations assigned to a cluster were classified as the semantic core, whereas those labeled as noise were classified as the tail. The core fraction was defined as the proportion of explanations assigned to a cluster within each condition. Sensitivity analyses varying parameter *min_cluster_size* from two to 10 yielded the same qualitative human–GenAI pattern.

**Theory revision.** We measured the extent of theory revision as the cosine distance between each contributor's pre- and post-ML explanation embeddings:

$$d_i = 1 - \frac{\boldsymbol{z}_{i,\text{pre}} \cdot \boldsymbol{z}_{i,\text{post}}}{\| \boldsymbol{z}_{i,\text{pre}} \|_2 \| \boldsymbol{z}_{i,\text{post}} \|_2}.$$

Here, $\boldsymbol{z}_{i,\text{pre}}$ and $\boldsymbol{z}_{i,\text{post}}$ denote contributor $i$'s pre- and post-ML explanation embeddings. Larger values indicated greater semantic change between the two phases.

Human–GenAI differences in revision magnitude were evaluated using two-sided Welch's $t$-tests. Associations between pre-ML prediction accuracy and revision magnitude were assessed using Spearman rank correlations and bivariate OLS regressions.

**Interactive visualization.** An interactive visualization of the semantic networks of theoretical explanations is available at https://kelichloe.github.io/human-ai-semantic-network/index.html.

### M.5 Theory evaluation

Theory quality was evaluated independently by two expert human raters and an LLM evaluator. The human raters were Full Professors with expertise in racial and gender inequality and human–AI interaction, respectively. The LLM evaluator was GPT-5.5 with high reasoning effort. All theories were anonymized, randomly ordered, and evaluated blind to contributor type and pre- versus post-ML status. The LLM rated the full corpus of 784 theories (98 contributors × 2 tasks × 2 effect types × 2 phases), whereas the human raters evaluated stratified weighted random subsets of 160 and 80 theories, respectively. Overall theory quality was the mean of scores on five dimensions: clarity and coherence, causal reasoning, theoretical depth, creativity, and persuasiveness. The complete rating rubric for the human and LLM evaluators is provided in the Supplementary Information.

Agreement between human-produced and LLM-produced ratings was assessed using Pearson and Spearman correlations with two-sided *P* values. Mean ratings are reported with 95% bootstrap confidence intervals based on 5,000 resamples. Between-group differences and pre- versus post-ML differences were tested using two-sided Welch's *t*-tests.

**M.6 Data availability**

The dataset used for the machine-learning analyses of racial- and gender-inequality discourse is publicly available through Zenodo at https://doi.org/10.5281/zenodo.22647761. The human and GenAI survey data underlying the analyses reported in this study are publicly available in the GitHub repository associated with this study at https://github.com/KeLiChloe/human-ai-theory-building.

**M.7 Code availability**

All code used for the analyses reported in this study is publicly available at https://github.com/KeLiChloe/human-ai-theory-building. The repository contains code for the analysis of the human and GenAI survey data and for the machine-learning analyses used to construct the empirical benchmarks. Detailed instructions for reproducing the analyses are also provided in the repository.

# Supplementary Information for Li et al. (2026)

# “Artificial intelligences and human scientists exhibit complementary strengths in theory building”

## Table of Contents

## S.1 Deviations from the preregistered analysis plan

The study was preregistered on the Open Science Framework (OSF) before collecting data: https://osf.io/buah6/overview?view_only=98ca17155c4d4463acb6f7c043ba54b1. The preregistered analysis plan covered prediction performance, mechanistic complexity, the structure and diversity of theoretical explanations, blind theory-quality evaluation, and moderator analyses. The final manuscript also includes several further analyses that were not originally planned, as well as several refinements to and deviations from the analyses that were planned. We summarize these below.

**Refinements and extensions of preregistered analyses.** The preregistration specified broad approaches for characterizing prediction accuracy, mechanistic complexity and theory structure. In the final analyses, these constructs were operationalized using more specific measures. Prediction accuracy was summarized using signed cosine similarity, jointly capturing feature selection and predicted direction. Theory complexity was assessed from causal diagrams, using the number of causal paths, maximum path length and number of latent variables; and from textual theoretical explanations, using the word count. Theory diversity was operationalized using embedding-based centroid distance, mean pairwise cosine distance and HDBSCAN-based core–tail structure. These analyses address the same substantive questions specified in the preregistration but use more detailed final implementations. Also, the preregistered theory quality evaluation was broadened in scope. The preregistration specified blind human ratings of post-ML theories from senior experts and GenAI systems. In the final analysis, we included both pre- and post-ML theories, extended the evaluation to include PhD-student theories, and added an LLM evaluator using the same rating rubric given the large volume of contributor-generated theories to be evaluated. Moderator analyses were preregistered for doctoral students and focused on topic specialization and gender; the final analysis included all human contributors and jointly modeled gender, topic expertise, and seniority (in the last case, comparing professors and doctoral students).

**Additional analyses.** We conducted several statistical analyses that were not specified in the preregistration. These included crowd-level aggregation and aggregation gain; prediction and error diversity and their association with aggregation gain; analyses of prediction error patterns; the association between prior prediction accuracy and subsequent theory revision; and analyses relating theory complexity to prediction accuracy and theory-quality ratings.

**Changes to the originally planned analyses.** There were two main changes to the preregistered analyses. First, for prediction accuracy, the preregistration proposed separately assessing feature overlap, rank correspondence with ML importance and sign agreement. The final analysis instead used signed cosine similarity as a unified measure of alignment between contributor predictions and the ML benchmark. Second, for mechanistic complexity, the preregistration proposed separate coding of narrative explanations and causal diagrams. The current manuscript reports the diagram-based complexity analyses but does not include a separate textual complexity analysis, other than a word-count analysis.

**Preregistered analysis not reported here.** The analysis plan originally included evaluating the empirical support of contributor-generated theory–test pairs. We did not conduct or report these analyses in the present paper because they will be the topic of a follow-up paper.

## S.2 Machine-learning empirical benchmarks

### S.2.1 Dictionaries of keywords for labeling papers

- **Gender Inequality Keywords (63 words):**
  - *"marginalize women", "marginalization of women", "sexual discrimination", "sex discrimination", "gender discrimination", "gender stereotypes", "gender stereotyping", "sex stereotypes", "sex stereotyping", "male dominance", "male dominated", "female subordination", "gendered division of labor", "gender wage gap", "gender wage gaps", "gender bias", "gender biases", "female underrepresentation", "underrepresentation of female", "women underrepresentation", "underrepresentation of women", "gender disparity", "gender disparities", "sex disparity", "sex disparities", "gender norm", "gender norms", "sexism", "sexist", "sexists", "patriarchy", "sex bias", "sex biases", "gender pay gap", "gender pay gaps", "gender income gap", "gender income gaps", "gendered expectation", "gendered expectations", "gender-based violence", "objectification", "pink tax", "mansplaining", "manterrupting", "gender-based harassment", "gender identity", "women's rights", "gender diversity", "gender balance", "feminism", "women empowerment", "feminist", "LGBTQ", "gender equality", "equal pay", "equal rights", "gender", "genders", "sex roles", "gender roles", "female and male", "women and men", "sexes".*

- **Racial Inequality Keywords (57 words):**
  - *"racial bias", "racial biases", "racial inequality", "racial inequalities", "racial injustice", "racial discrimination", "racial stigma", "racial wealth gap", "racial wealth gaps", "racial disparity", "racial disparities", "racial exclusion", "racial exclusions", "racism", "racial income", "racial oppression", "racial privilege", "racial privileges", "racial profiling", "racial stereotype", "racial stereotypes", "racial stereotyping", "racial hierarchy", "racial violence", "racial minority", "racial tension", "racial segregation", "racial intolerance", "racial scapegoating", "prejudice", "colorism", "white privilege", "xenophobia", "hate speech", "hate crimes", "racial microaggression", "racial microaggressions", "racial diversity", "racial equality", "racial equalities", "racial justice", "minority representation", "representation of minority", "racial equity", "anti-racism", "antiracist", "racial inclusion", "ethnicity", "ethnic groups", "race", "ethnic identity", "racial identity", "racial", "people of color", "communities of color", "racial groups", "racial demographics".*

### S.2.2 List of features

| Feature | Description and Construction |
|---|---|
| *social_science* | Equals 1 if the paper was published in a journal whose scope lies within the social sciences (e.g., sociology, political science, psychology, economics, history, or philosophy), and 0 otherwise. |
| *natural_science* | Equals 1 if the paper was published in a journal whose scope lies within the natural sciences (e.g., biology, chemistry, physics, environmental science, or medicine), and 0 otherwise. |
| *engineering_and_technology* | Equals 1 if the paper was published in a journal whose scope lies within engineering and technology (e.g., engineering, computer science, or applied mathematics), and 0 otherwise. |
| *num_authors* | Total number of authors of the paper. |
| *female* | Estimated share of female authors on the author team. For each co-author, we infer the probability of being female from their first name and average these probabilities across all co-authors. |
| *asian* | Estimated share of Asian authors on the author team. For each author, we infer the most likely country of birth from their last name. Based on the racial composition of that country, we estimate the probability that the author is Asian. The feature is the average of these estimated probabilities across all co-authors. |
| *black* | Estimated share of Black authors on the author team. For each author, we infer the most likely country of birth from their last name. Based on the racial composition of that country, we estimate the probability that the author is Black. The feature is the average of these estimated probabilities across all co-authors. |
| *hispanic* | Estimated share of Hispanic authors on the author team. For each author, we infer the most likely country of birth from their last name. Based on the racial composition of that country, we estimate the probability that the author is Hispanic. The feature is the average of these estimated probabilities across all co-authors. |
| *white* | Estimated share of White authors on the author team. For each author, we infer the most likely country of birth from their last name. Based on the racial composition of that country, we estimate the probability that the author is White. The feature is the average of these estimated probabilities across all co-authors. |

| Feature | Description and Construction |
|---|---|
| *authors_race_diversity_score* | Average racial diversity within the co-author team, measured by Shannon entropy. For each author, we first infer the most likely country of birth from their last name and assign a four-dimensional race-composition probability vector (asian, black, hispanic, white). These vectors are averaged across all co-authors to obtain a team-level racial distribution, and Shannon entropy is computed on the resulting distribution. Higher values indicate a more balanced racial composition within the author team, whereas lower values indicate dominance by a single racial group. |
| *country_race_diversity_score* | Average racial diversity of the authors' inferred countries of birth, measured by Shannon entropy. For each author, we first infer the most likely country of birth from their last name and assign a four-dimensional race-composition probability vector (asian, black, hispanic, white). Shannon entropy is computed for each author's probability vector and then averaged across all co-authors. Higher values indicate that authors tend to come from countries with more racially balanced populations. |
| *news_inequality_mentions_3_years* | Average annual percentage of news records mentioning inequality during the three years preceding the paper's publication year. |
| *paper_inequality_mentions_3_years* | Average annual percentage of academic paper records mentioning inequality during the three years preceding the paper's publication year. |

***Note***: the last two time-lagged inequality-attention variables were constructed using the databases built by Professor Philip M. Parker's data team at INSEAD. The news database contains approximately 700 million records. For each publication year, we calculated the proportion of records in each database that mentioned inequality-related terms, including gender, racial, economic, and general inequality. The variable *news_inequality_mentions_3_years* was defined as the average annual proportion of news records mentioning inequality during the three years preceding the focal paper's publication year. The variable *paper_inequality_mentions_3_years* was constructed analogously using the Semantic Scholar corpus. In this study, inequality-related terms appeared in approximately 0.7% of news records annually and in approximately 3–6% of academic publication records annually.

### S.2.3 Model specifications

For the random-forest ranking procedure, hyperparameters were selected by grid search with three-fold cross-validation, optimizing the macro-averaged F1 score. The tuning grid varied the number of trees (200, 400), maximum tree depth (12, 16, 20), minimum number of samples required to split an internal node (5, 10), minimum number of samples required at a leaf node (3, 5), and the number of features considered at each split (sqrt, 0.8). In each subsampling repetition, the data were randomly reshuffled and partitioned into ten non-overlapping, approximately equal-sized subsets, with a separate random forest fitted to each subset; this procedure was repeated 100 times. Within each fitted model, predictors were ranked by mean decrease in impurity (MDI). A predictor received one vote whenever it appeared among the five highest-ranked features in each run, and predictors were ordered by their total vote count, with ties broken by mean MDI. For logistic regression, hyperparameters were selected by grid search with stratified five-fold cross-validation, optimizing ROC-AUC. The tuning grid varied the solver (liblinear, lbfgs), maximum number of iterations (500, 1,000), and inverse regularization strength $C$ (0.01, 0.1, 0.3, 1, 3, 10). The direction of each selected predictor or interaction was determined from the sign of its mean coefficient across runs. We constructed 95% confidence intervals around the mean coefficient using the standard error across runs and the corresponding $t$-distribution critical value. Random forest models, logistic regression models, and cross-validated hyperparameter tuning were implemented in scikit-learn.

## S.3 Study contributors

### S.3.1 Human sample demographics

We initially recruited 14 faculty members and 61 doctoral students, but one member of each group was removed for completing the survey too quickly (in 10-15 minutes or less, when the other contributors took hours) and for articulating extremely brief and under-elaborated theories (e.g., a theoretical statement with only two words in length). Thus, our final sample consisted of 13 senior and 60 junior academics. Demographic characteristics are reported separately for the two groups below.

- **PhD Students (n = 60)**

| | n | % |
|---|---|---|
| ***Gender*** | | |
| Male | 40 | 66.7 |
| Female | 18 | 30.0 |
| Non-binary | 1 | 1.7 |
| Prefer not to disclose | 1 | 1.7 |
| ***Country of residence*** | | |
| United States | 45 | 75.0 |
| France | 5 | 8.3 |
| Singapore | 5 | 8.3 |
| United Kingdom | 3 | 5.0 |
| Canada | 1 | 1.7 |
| Switzerland | 1 | 1.7 |
| ***Topic expertise*** | | |
| Topic expert | 6 | 10.0 |
| Non-expert | 54 | 90.0 |

| | n | % |
|---|---|---|
| ***Discipline*** | | |
| Operations Management | 11 | 18.3 |
| Computing & Information Systems | 10 | 16.7 |
| Management & Strategy | 9 | 15.0 |
| Statistics & Data Science | 8 | 13.3 |
| Engineering & Physical Sciences | 5 | 8.3 |
| Marketing & Economics | 5 | 8.3 |
| Psychology & Organizational Behavior | 5 | 8.3 |
| Decision Sciences | 3 | 5.0 |
| Mathematics | 3 | 5.0 |
| Life Sciences | 1 | 1.7 |
| ***Institution*** | | |
| INSEAD | 11 | 18.3 |
| Massachusetts Institute of Technology | 11 | 18.3 |
| University of Pennsylvania | 9 | 15.0 |
| Cornell University | 6 | 10.0 |
| New York University | 4 | 6.7 |
| Columbia University | 2 | 3.3 |
| Johns Hopkins University | 2 | 3.3 |
| Northwestern University | 2 | 3.3 |
| Stanford University | 2 | 3.3 |
| University of California, Berkeley | 2 | 3.3 |
| University of Cambridge | 2 | 3.3 |
| ETH Zurich | 1 | 1.7 |
| Indiana University | 1 | 1.7 |
| London Business School | 1 | 1.7 |
| Stockholm School of Economics | 1 | 1.7 |
| UC Berkeley & UCSF | 1 | 1.7 |
| University of Chicago | 1 | 1.7 |
| University of Toronto | 1 | 1.7 |

- **Senior Scientists / Professors (n = 13)**

| | n | % |
|---|---|---|
| ***Gender*** | | |
| Male | 8 | 61.5 |
| Female | 5 | 38.5 |
| ***Country of residence*** | | |
| United States | 5 | 38.5 |
| Canada | 2 | 15.4 |
| Singapore | 2 | 15.4 |
| United Kingdom | 2 | 15.4 |
| France | 1 | 7.7 |
| Israel | 1 | 7.7 |
| ***Topic expertise*** | | |
| Topic expert | 11 | 84.6 |
| Non-expert | 2 | 15.4 |
| ***Discipline*** | | |
| Psychology & Organizational Behavior | 10 | 76.9 |
| Management & Strategy | 3 | 23.1 |
| ***Institution*** | | |
| ESSEC Business School | 1 | 7.7 |
| INSEAD | 1 | 7.7 |
| London Business School | 1 | 7.7 |
| New College of Florida | 1 | 7.7 |
| Oregon State University | 1 | 7.7 |
| Reichman University | 1 | 7.7 |
| Singapore Management University | 1 | 7.7 |
| Stanford University | 1 | 7.7 |
| University College London | 1 | 7.7 |
| University of North Texas | 1 | 7.7 |
| University of Pennsylvania | 1 | 7.7 |
| University of Toronto | 1 | 7.7 |
| University of Waterloo | 1 | 7.7 |

### S.3.2 GenAI model contributors

The GenAI contributor group consisted of 25 large language model configurations that completed the same survey tasks as the human contributors.

- **Claude:** *Claude opus4.7, Claude opus4.7 adaptive, Claude opus4.8, Claude opus4.8 adaptive, Claude sonnet4.6*
- **ChatGPT**: *GPT 5.2 instant, GPT 5.3 instant, GPT 5.4 thinking extended, GPT 5.4 thinking standard, GPT 5.5 thinking extended, GPT 5.5 thinking standard, GPT 5.6 Sol high, Theorista GPT5.5 instant, Theorista GPT5.5 thinking standard* (Note. Theorista is a custom GPT developed by Professor Phanish Puranam as an AI thinking partner to help construct, derive, and test causal theories in the social sciences.)
- **Gemini**: *Gemini 3 fast, gemini 3.1 pro standard, gemini 3.6flash extended*
- **Qwen**: *Qwen 3.6Plus thinking, Qwen 3.7Plus thinking, Qwen 3.8Max thinking*
- **DeepSeek**: *DeepSeek expert-V4*
- **Grok:** *Grok4 expert, Grok4 fast*
- **Mistral:** *Mistral balanced, Mistral thinking*

### S.3.3 Potential exposure of GenAI systems to the underlying literature

Because the corpus included papers published through 2021, some GenAI systems may have encountered portions of the underlying scholarly literature during pretraining. However, the prediction task did not ask models to retrieve specific papers or previously reported findings. Models were provided with the candidate feature definitions and asked to predict which main effects and pairwise interactions would rank as most predictive in our independently constructed machine-learning benchmark, together with the expected direction of those associations. The benchmark itself was generated from our study-specific corpus, labeling procedure, feature construction, robust ML analyses, and was not shown to the LLM models before they completed the prediction task. Models were accessed through their standard conversational interfaces and were not allowed to use browsing or external tools during the study. Thus, prior exposure to individual papers does not constitute direct leakage of the benchmark, although we cannot rule out indirect familiarity with broader scholarly regularities represented in the corpus.

## S.4 LLM prompts used in data analysis

### S.4.1 Extraction of theory-complexity measures for diagrams

**Model:** gpt-5.5
**Reasoning effort:** high

**System prompt:**
You are a careful research assistant helping code open-ended survey responses.

---

PROJECT CONTEXT

---

The project studies human-AI collaboration in scientific theory building. Respondents forecast whether academic papers discuss inequality-related topics, especially racial inequality and gender inequality. The survey asks respondents to construct theories about predictors of whether a paper discusses racial inequality.

---

DATASET CONTEXT

---

Each paper has 13 observed features. These are the only official observed features in the project. See attached file for details. The outcome is whether a paper discusses racial inequality. Respondents may refer to the outcome as: Y, racial inequality paper, race inequality research, probability of race inequality research, likelihood of racial inequality publication, or similar.

---

SURVEY QUESTION

---

Respondents were instructed: "Please provide a diagram with arrows, expressed in text form, to represent your theory, where an arrow indicates a causal relationship between two variables. You may use the arrow symbol '→' to represent the causal direction. Please also indicate the sign of the causal effect between the two variables connected by each arrow, e.g. '+' or '−'."

---

YOUR CODING TASK

---

For each respondent's response, code exactly three quantities:

1. number of paths

Count the number of distinct causal paths explicitly written by the respondent. A path is usually one causal chain separated by arrows, often appearing as one line or one sentence. Example: A→B→Y C→Y counts as 2 paths.

2. maximum path length

Count the maximum number of causal arrows in any one path. Example: A → B → C → Y has length 3. A → Y has length 1. If no causal arrow or causal relation is present, use 0.

3. number of latent variables

Count the number of unique variables/concepts in the diagram that are NOT one of the 13 official observed features and are NOT the outcome Y. These include mediators, mechanisms, latent constructs, invented concepts, or renamed theoretical constructs. Examples: topic fit, topic popularity, interest, awareness, comfort, salience, academic trend, perceived legitimacy. Do not count:

Do not count:

• Y or any synonym of the racial inequality outcome

• the 13 official observed features, even if written with minor wording variations

• signs such as + or −

• generic labels such as X1, X2, mediator, Med1, if they merely label a substantive variable already named nearby

————————————————

IMPORTANT NORMALIZATION RULES

————————————————

Treat common natural-language variants as equivalent to the official features:
• social science, social sciences, social science, or similar → social_science
• natural science, natural sciences, natural science, or similar → natural_science
• engineering, technology, computer science, engineering and technology, or similar → engineering_and_technology
• number of authors, team size, or similar → num_authors
• female authors, share of female authors, female score, or similar → female
• Asian authors, or similar → asian
• Black authors, or similar → black
• Hispanic authors, other race authors → hispanic and other
• White authors, majority white team, or similar → white
• author race diversity, author racial diversity, racially diverse author team, or similar → authors_race_diversity_score
• country race diversity, racially diverse countries, country diversity, or similar → country_race_diversity_score
• news inequality mentions, media attention, news attention to inequality, or similar → news_inequality_mentions_3_years
• paper inequality mentions, prior publications, academic attention, recent academic attention to inequality, or similar → paper_inequality_mentions_3_years

Be conservative but thoughtful. The responses are unstructured and may contain prose, arrows, parentheses, signs, line breaks, and inconsistent names. Use careful reasoning to infer the diagram structure.

Return only valid JSON matching the requested schema.

**User prompt:**

Please analyze the following diagram response. Respondent response: """ {contributor_response_text} """

Return:

1. number of paths

> 2. maximum path length
> 3. number of latent variables
> 4. brief reasoning

**Human validation of diagram coding.** To assess the reliability of the LLM coding, a human independently coded a stratified random sample of diagrams (five each from racial- and gender-inequality tasks in the pre-ML and post-ML phases), using the same coding rule. The human coder was blind to the LLM codes during the process. Initial exact agreement with the LLM was 18/20 for number of paths, 19/20 for maximum path length, and 17/20 for number of latent variables. We then adjudicated the disagreements, which occurred in four diagrams:

- In one diagram the human coder judged the diagram unreadable/uncodeable, whereas the LLM assigned numeric codes.
- In a second diagram the human coder omitted two mediated paths, undercounting the number of paths by two; after correction, the human codes matched the LLM.
- In the remaining two diagrams the only discrepancy was the number of latent variables; upon review, the LLM counts were judged correct.

Thus, after post-review, human agreement with the LLM was 100% among diagrams judged codable by the human coder.

### S.4.2 LLM theory evaluation

Evaluations were run separately for each domain (racial inequality, gender inequality), and effect type (main effects, second-order interactions). Within each condition, explanations were first anonymized with stable theory identifiers, randomly shuffled (seed 20260811), and then scored in batches of up to 25 theories per API call. For each batch, the model received a fixed system prompt describing the prediction task, the available features, and five evaluation dimensions, together with a user message that listed the anonymized theory texts. The required response format was a structured JSON object containing, for every theory identifier in the batch, integer scores on a 1–10 scale for each of the five dimensions, plus a short free-text justification (brief reasoning). Responses that omitted, duplicated, or invented theory identifiers were rejected and retried. Overall theory quality was defined post hoc as the unweighted mean of the scores on the five dimensions.

> **Model:** gpt-5.5
> **Reasoning effort:** high
>
> **System prompt:**
> You are an expert in social science research and theory evaluation. Use your strongest analytical judgment and highest level of social-scientific reasoning. Be rigorous, discerning, and careful.
>
> ────────────────────────────
>
> PROJECT CONTEXT
>
> ────────────────────────────

> This study examines theory building for predicting mentions of {domain} in academic papers.
> Available features: [feature list omitted; see Section S.2.2]
>
> ———————————————
>
> TASK
>
> ———————————————
>
> You will receive a batch of theoretical explanations. Each item has an anonymous theory id and the theory text. Evaluate each theory independently on its own merits.
>
> {task_blurb}
>
> ———————————————
>
> EVALUATION DIMENSIONS (1–10 scale)
>
> ———————————————
>
> For each dimension, assign a score from 1 (very poor) to 10 (excellent).
> 1. Clarity and Coherence Is the explanation clearly written, well-structured, and logically consistent, without ambiguity or internal contradictions?
> 2. Causal Reasoning Does the explanation articulate plausible causal mechanisms relevant to the outcome?
> 3. Theoretical Depth Does the explanation go beyond surface-level statements and engage with meaningful underlying concepts or mechanisms?
> 4. Creativity Does the explanation demonstrate creative or original thinking, such as offering novel perspectives, non-obvious connections, or insightful interpretations?
> 5. Persuasiveness Does the explanation provide a convincing theoretical account?
>
> ———————————————
>
> SCORING GUIDELINES
>
> ———————————————
>
> 1–2 = poor 3–4 = weak 5–6 = moderate 7–8 = strong 9–10 = excellent
>
> ———————————————
>
> OUTPUT
>
> ———————————————
>
> Return scores for EVERY theory id in the batch. For each theory, provide the five-dimension scores and a brief reasoning (at most 5 sentences) that justifies the scores. Do not omit any theory id. Do not invent theory ids that were not provided.
>
> **User prompt:**
> Evaluate each of the following theoretical explanations independently. Theory IDs in this batch (in order presented): {ids}.
> Return one score object per theory id listed above.

**Task blurbs (task_blurb):**

**Main effects:** *"Each respondent selected 5 features from the list below (with a predicted positive or negative association with the outcome) and wrote a theoretical explanation."*

**Second-order interactions:** *“Each respondent selected 3 two-way interactions among the features below (second-order interactions / SOI, each with a predicted positive or negative association with the outcome) and wrote a theoretical explanation.”*

## S.5 Survey contents

We used a Qualtrics survey to invite contributors to the theory-building task. Below, we present the survey content in text format.

----------------- START OF SURVEY -----------------

**Human-AI Collaboration - Forecasting Survey on Inequality Discourses**

**What is your full name?**

**What is your email address?**

**Authorship**

In recognition of your contribution, colleagues who complete the survey and meet the project requirements will be included as co-authors on the first publication resulting from this forecasting study.

**Instructions**

Important: Please do not use any AI tools during the survey, as this would compromise the integrity of the study. The survey is expected to take approximately 1–2 hours. We kindly ask that you set aside sufficient uninterrupted time to complete it thoughtfully. For the best experience, we recommend completing this survey on a PC or laptop, and keeping a pen and paper nearby in case you wish to take notes. If you have any questions, please feel free to contact Ke Li (ke.li@insead.edu).

**Consent Statement**

By signing below, I confirm that I will not use any AI tools or external models while completing this survey, as I understand that doing so would compromise the integrity of the research.

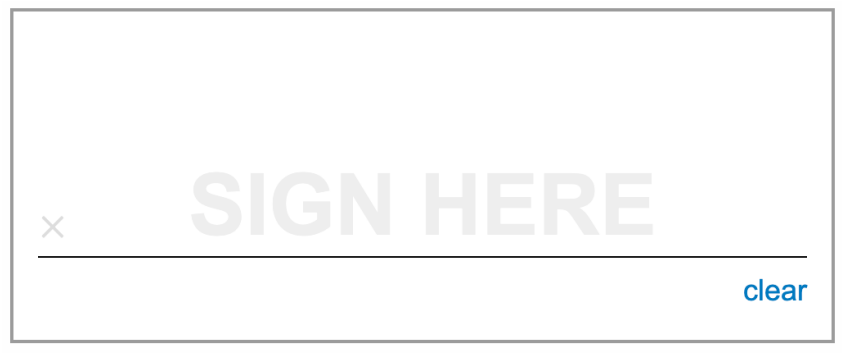

**Introduction**

This survey is part of a broader research project that develops a human–AI collaborative framework for theory building. We investigate how different forms of intelligence—such as human judgment, machine learning, and generative AI—may complement one another in the scientific process. In terms of the specific topic, we examine how academic discourses on social inequality have evolved over time (e.g., from the 2000s to the 2020s) and across different social groups (e.g., racial and ethnic groups, women vs. men). Using large-scale textual data, we identify empirical patterns in who discusses what forms of social inequality. In this survey, we ask you to draw on your theoretical knowledge and causal reasoning to predict patterns in the data and to propose explanatory theories. Your responses will contribute to a broader exploration of how humans and AI systems can work together to advance theory development.

**Dataset Description**

Our study is based on the Semantic Scholar Academic Corpus, which contains published papers across a wide range of disciplines, including psychology, social sciences, engineering, economics, mathematics, computer science, and others. We sampled approximately 150,000 papers to construct our dataset.

To identify academic discussions related to social inequality, we previously invited expert scholars in this area to construct a list of inequality-related keywords. Examples are provided below (the full list is available here).

- Racial Inequality: *racial bias, racial diversity, racial equality, racial inequality, hate crime, racial discrimination, racial justice, racial violence, racial stereotypes*
- Gender Inequality: *patriarchy, male dominance, gender identity, gender stereotypes, sexism, gender biases, sex discrimination*

We label each paper according to the following rules:

- Positive sample (label = "1"): The title or abstract of the paper contains at least one of the inequality-related keywords.
- Negative sample (label = "0"): The title or abstract of the paper contains none of these inequality-related keywords.

**Overview of Your Tasks**

You will complete two sets of tasks, with each set containing 17 questions:

- Predicting mentions of **racial** inequality in academic papers
- Predicting mentions of **gender** inequality in academic papers

**Predicting Mentions of Racial Inequality - Phase 1: Human Forecasting**

In this phase, we aim to capture your independent human reasoning, without prior knowledge of the empirical patterns identified by our machine-learning analysis. In this task, the goal is to predict whether a paper discusses **racial** inequality. A paper is classified as discussing **racial** inequality if its title or abstract contains at least one **racial** inequality keyword.

- Examples of **racial** inequality keywords: *racial bias, racial diversity, racial equality, racial inequality, hate crime, racial discrimination, racial justice, racial violence, racial stereotypes*

We have collected 13 features for each paper. You may refer to the tables below or to this feature list, for detailed definitions of each feature.

Here, the survey displayed the 13 features and their definitions. See Section S.2.2 ("List of Features").

**Q Race.1**

Select the top five features that you believe are most important for this **racial** inequality prediction task. Here, "important" refers to features that you believe are most strongly associated with whether a paper discusses **racial** inequality, either positively or negatively associated.

Here, the survey displayed the names of the 13 features listed in Section S.2.2, with a checkbox next to each. The Qualtrics survey was configured to require the selection of exactly five features.

**Q Race.2**

Please drag and rank your selected features from most to least important for the **racial** inequality task. Here, "important" refers to features that you believe are most strongly associated with whether a paper discusses **racial** inequality, either positively or negatively associated.

Here, the survey displayed the five features selected in Q Race.1.

**Q Race.3**

Please indicate the sign of the effect of each chosen feature.

- Positive sign (+): A positive sign (+) means that you believe higher values of this feature are associated with a higher probability that a paper discusses **racial** inequality.
- Negative sign (–): A negative sign (−) means that you believe higher values of this feature are associated with a lower probability that a paper discusses **racial** inequality.

Here, the survey displayed the five features selected in Q Race.1, with a positive (+) and negative (−) option next to each.

**Q Race.4**

Please provide a brief theoretical explanation about why these features are most important, their rank, and the signs you indicated. Explain the assumptions or reasoning behind your beliefs. Your answers should be a narrative paragraph in English.

Here, the survey displayed a text input box

**Q Race.5**

Please provide a diagram with arrows, expressed in text form, to represent your theory, where an arrow indicates a causal relationship between two variables.

- You may use the arrow symbol "→" to represent the causal direction (you can copy the symbol "→" directly).
- Please also indicate the sign of the causal effect between the two variables connected by each arrow (e.g., "+" for a positive effect and "–" for a negative effect).

See below example.

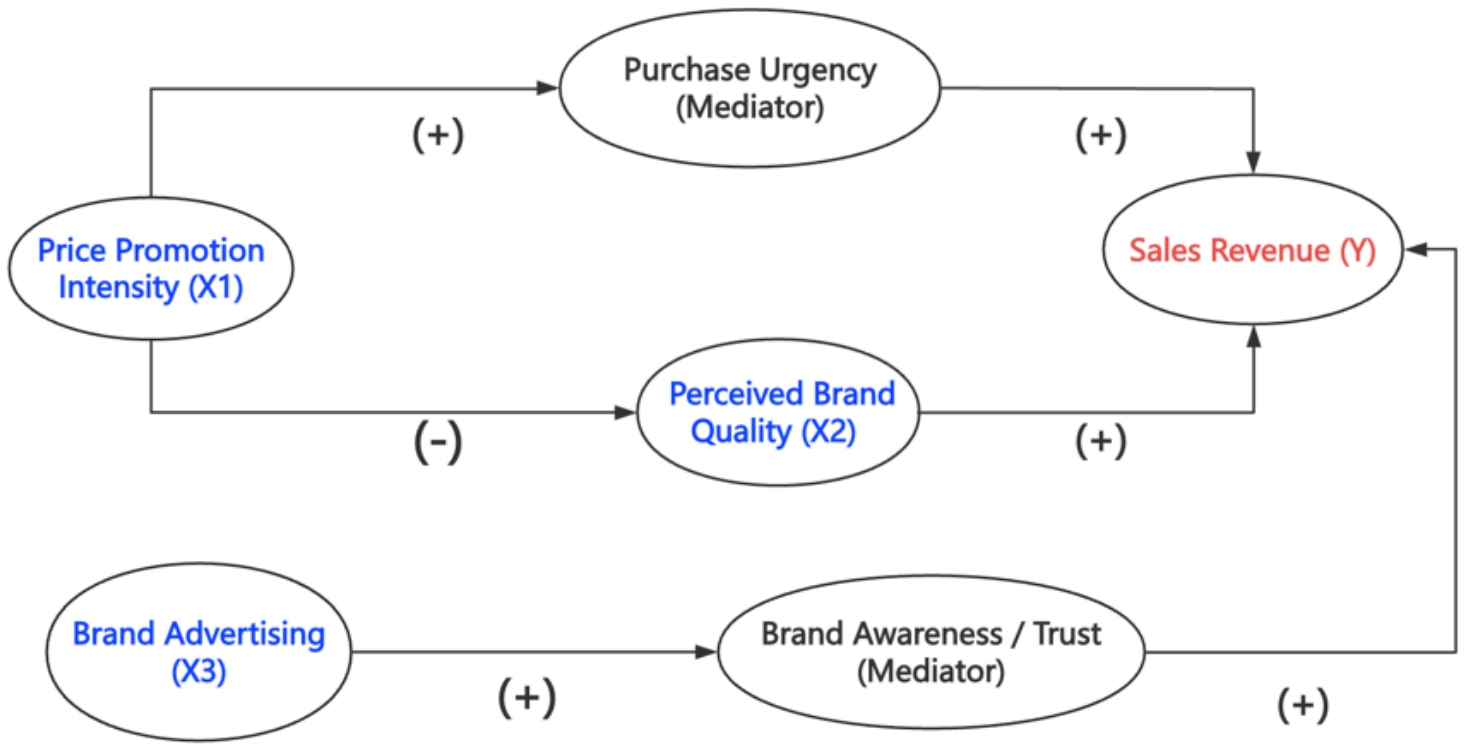

For example, suppose you model the effects of price promotion intensity (X1), perceived brand quality (X2), and brand advertising (X3) on sales revenue (Y) using the diagram shown above. You should express all the distinct paths as follows:

*Price Promotion Intensity (X1) → (+) Purchase Urgency (mediator) → (+) Sales Revenue (Y)*

*Price Promotion Intensity (X1) → (–) Perceived Brand Quality (X2) → (+) Sales Revenue (Y)*

*Brand Advertising (X3) → (+) Brand Awareness / Trust (mediator) → (+) Sales Revenue (Y)*

Here, the survey displayed a text input box

**The following questions focus on second-order interactions.** What Are Second-Order Interactions? Second-order interactions describe how two independent variables ($X_1$ and $X_2$) jointly influence an outcome ($Y$), beyond their separate effects. In other words, the impact of one predictor depends on the value of another predictor. A common way to model this is by including a product term $X_1 * X_2$ in the regression model. The sign of an interaction shows whether the combined effect enhances or diminishes the outcome:

- Positive interaction (+): The marginal effect of one variable on $Y$ becomes stronger as the other increases—they *reinforce* each other's influence.
- Negative interaction (−): The marginal effect of one variable on $Y$ weakens as the other increases—they *offset* each other's influence.

Example: Suppose a company test how Discounts ($D$) and Ad Spending ($A$) affect Sales ($S$) using the regression model: $S = \beta_0 + \beta_1 D + \beta_2 A + \beta_3 (D \times A) + \varepsilon$

Here, $D \times A$ represents the second-order interaction between Discounts and Ad Spending.

- Positive interaction ($\beta_3 > 0$): The marginal effect of discounts on sales increases with ad spending, and the marginal effect of ad spending increases with discounts — the two strategies reinforce each other to boost sales.
- Negative interaction ($\beta_3 < 0$): The marginal effect of discounts on sales decreases with ad spending, and the marginal effect of ad spending decreases with discounts — the two strategies weaken each other's incremental impact on sales.

**Q Race.6**

Choose the second-order interaction that you believe is <u>the most</u> important for predicting whether a paper discusses **racial** inequality or not, by selecting two features. Here, "important" refers to interactions that you believe are most strongly associated with whether a paper discusses **racial** inequality, either positively or negatively associated.

> Here, the survey displayed the 13 features, with a checkbox next to each. Respondents were required to select exactly two features.

**Q Race.7**

Choose the second-order interaction that you believe is <u>the second most</u> important for predicting whether a paper discusses **racial** inequality or not, by selecting two features. Here, "important" refers to interactions that you believe are most strongly associated with whether a paper discusses **racial** inequality, either positively or negatively associated.

> Here, the survey displayed the 13 features, with a checkbox next to each. Respondents were required to select exactly two features.

**Q Race.8**

Choose the second-order interaction that you believe is the third most important for predicting whether a paper discusses **racial** inequality or not, by selecting two features. Here, "important" refers to interactions that you believe are most strongly associated with whether a paper discusses **racial** inequality, either positively or negatively associated.

Here, the survey displayed the 13 features, with a checkbox next to each. Respondents were required to select exactly two features.

**Q Race.9**

Now please indicate the signs of the interactions you've chosen.

- Positive sign (+): The effect of one predictor on Y (whether a paper discusses **racial** inequality) increases as the other predictor increases — meaning they reinforce each other's influence on Y.
- Negative sign (–): The effect of one predictor on Y (whether a paper discusses **racial** inequality) decreases as the other predictor increases — meaning they weaken each other's influence on Y.

Here, the survey displayed the three interactions selected in Q Race.6-8, with a positive (+) and negative (−) option next to each.

**Q Race.10**

Please provide a brief theoretical explanation in a narrative paragraph for your choices. Explain the assumptions or reasoning behind your beliefs about why these interactions are most important and why they take the signs you indicated.

Here, the survey displayed a text input box

**Predicting Mentions of Gender Inequality - Phase 1: Human Forecasting**

In this phase, we aim to capture your independent human reasoning, without prior knowledge of the empirical patterns identified by our machine-learning analysis. In this task, the goal is to predict whether a paper discusses **gender** inequality. A paper is classified as discussing **gender** inequality if its title or abstract contains at least one **gender** inequality keyword.

- Examples of **gender** inequality keywords: *patriarchy, male dominance, gender identity, gender stereotypes, sexism, gender biases, sex discrimination*

We have collected 13 features for each paper. You may refer to the tables below or to this feature list, for detailed definitions of each feature.

Here, the survey displayed the 13 features and their definitions. See Section S.2.2 ("List of Features").

### Q Gender.1

Select the top five features that you believe are most important for this **gender** inequality prediction task. Here, "important" refers to features that you believe are most strongly associated with whether a paper discusses **gender** inequality, either positively or negatively associated.

Here, the survey displayed the names of the 13 features listed in Section S.2.2, with a checkbox next to each. The Qualtrics survey was configured to require the selection of exactly five features.

### Q Gender.2

Please drag and rank your selected features from most to least important for the **gender** inequality task. Here, "important" refers to features that you believe are most strongly associated with whether a paper discusses **gender** inequality, either positively or negatively associated.

Here, the survey displayed the five features selected in Q Gender.1.

## Q Gender.3

Please indicate the sign of the effect of each chosen feature.

- Positive sign (+): A positive sign (+) means that you believe higher values of this feature are associated with a higher probability that a paper discusses **gender** inequality.
- Negative sign (–): A negative sign (−) means that you believe higher values of this feature are associated with a lower probability that a paper discusses **gender** inequality.

Here, the survey displayed the five features selected in Q Gender.1, with a positive (+) and negative (−) option next to each.

## Q Gender.4

Please provide a brief theoretical explanation about why these features are most important, their rank, and the signs you indicated. Explain the assumptions or reasoning behind your beliefs. Your answers should be a narrative paragraph in English.

Here, the survey displayed a text input box

## Q Gender.5

Please provide a diagram with arrows, expressed in text form, to represent your theory, where an arrow indicates a causal relationship between two variables.

- You may use the arrow symbol "→" to represent the causal direction (you can copy the symbol "→" directly).
- Please also indicate the sign of the causal effect between the two variables connected by each arrow (e.g., "+" for a positive effect and "–" for a negative effect).

See below example.

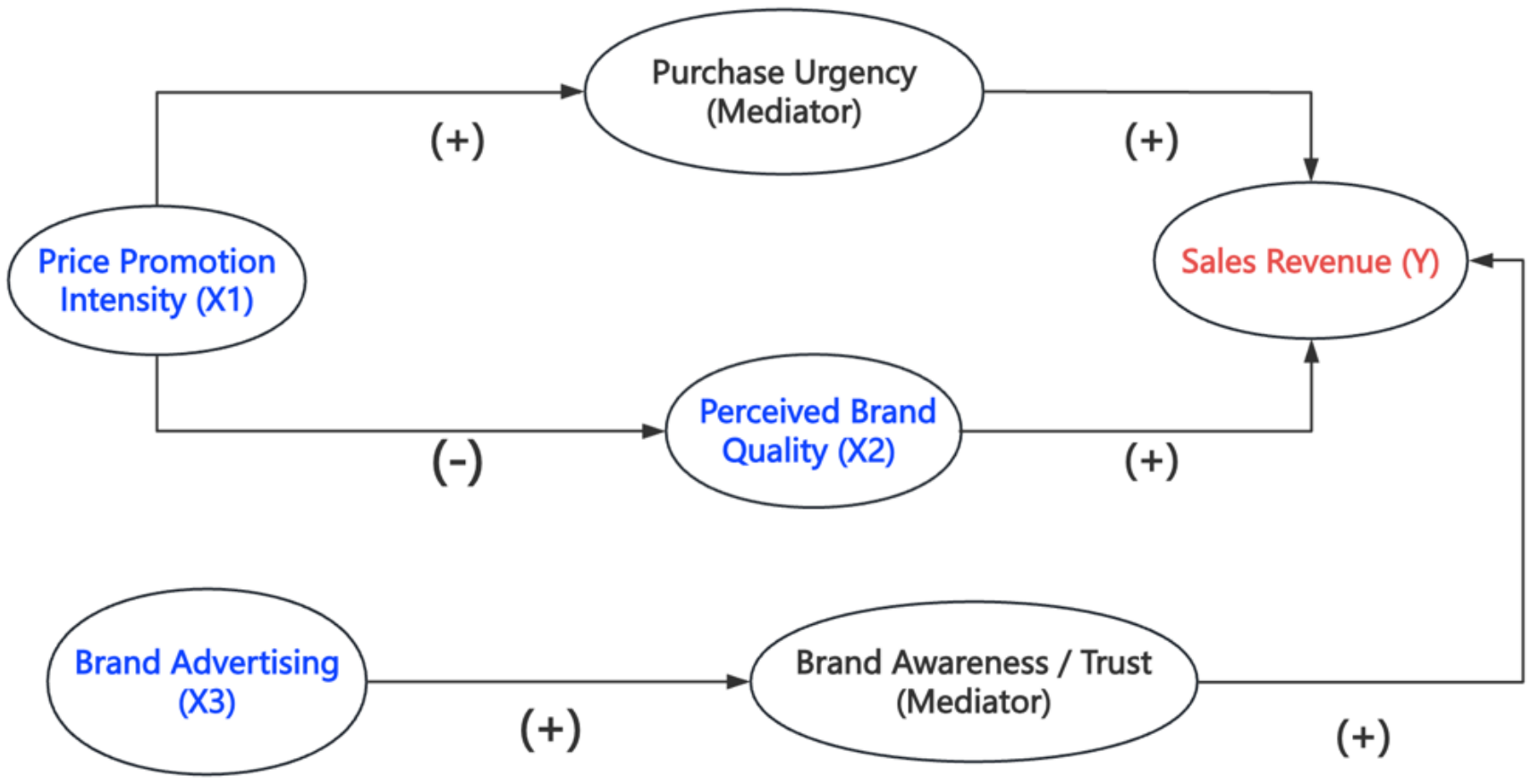

For example, suppose you model the effects of price promotion intensity (X1), perceived brand quality (X2), and brand advertising (X3) on sales revenue (Y) using the diagram shown above. You should express all the distinct paths as follows:

*Price Promotion Intensity (X1) → (+) Purchase Urgency (mediator) → (+) Sales Revenue (Y)*

*Price Promotion Intensity (X1) → (–) Perceived Brand Quality (X2) → (+) Sales Revenue (Y)*

*Brand Advertising (X3) → (+) Brand Awareness / Trust (mediator) → (+) Sales Revenue (Y)*

Here, the survey displayed a text input box

**The following questions focus on second-order interactions.** What Are Second-Order Interactions? Second-order interactions describe how two independent variables ($X_1$ and $X_2$) jointly influence an outcome ($Y$), beyond their separate effects. In other words, the impact of one predictor depends on the value of another predictor. A common way to model this is by including a product term $X_1 * X_2$ in the regression model. The sign of an interaction shows whether the combined effect enhances or diminishes the outcome:

- Positive interaction (+): The marginal effect of one variable on $Y$ becomes stronger as the other increases—they *reinforce* each other's influence.
- Negative interaction (−): The marginal effect of one variable on $Y$ weakens as the other increases—they *offset* each other's influence.

Example: Suppose a company test how Discounts ($D$) and Ad Spending ($A$) affect Sales ($S$) using the regression model: $S = \beta_0 + \beta_1 D + \beta_2 A + \beta_3 (D \times A) + \varepsilon$

Here, $D \times A$ represents the second-order interaction between Discounts and Ad Spending.

- Positive interaction ($\beta_3 > 0$): The marginal effect of discounts on sales increases with ad spending, and the marginal effect of ad spending increases with discounts — the two strategies reinforce each other to boost sales.
- Negative interaction ($\beta_3 < 0$): The marginal effect of discounts on sales decreases with ad spending, and the marginal effect of ad spending decreases with discounts — the two strategies weaken each other's incremental impact on sales.

**Q Gender.6**
Choose the second-order interaction that you believe is the most important for predicting whether a paper discusses **gender** inequality or not, by selecting two features. Here, "important" refers to interactions that you believe are most strongly associated with whether a paper discusses **gender** inequality, either positively or negatively associated.

Here, the survey displayed the 13 features, with a checkbox next to each. Respondents were required to select exactly two features.

**Q Gender.7**
Choose the second-order interaction that you believe is the second most important for predicting whether a paper discusses **gender** inequality or not, by selecting two features. Here, "important" refers to interactions that you believe are most strongly associated with whether a paper discusses **gender** inequality, either positively or negatively associated.

Here, the survey displayed the 13 features, with a checkbox next to each. Respondents were required to select exactly two features.

**Q Gender.8**
Choose the second-order interaction that you believe is the third most important for predicting whether a paper discusses **gender** inequality or not, by selecting two features. Here, "important" refers to interactions that you believe are most strongly associated with whether a paper discusses **gender** inequality, either positively or negatively associated.

Here, the survey displayed the 13 features, with a checkbox next to each. Respondents were required to select exactly two features.

**Q Gender.9**

Now please indicate the signs of the interactions you've chosen.

- Positive sign (+): The effect of one predictor on Y (whether a paper discusses **racial** inequality) increases as the other predictor increases — meaning they reinforce each other's influence on Y.
- Negative sign (–): The effect of one predictor on Y (whether a paper discusses **racial** inequality) decreases as the other predictor increases — meaning they weaken each other's influence on Y.

Here, the survey displayed the three interactions selected in Q Gender.6-8, with a positive (+) and negative (−) option next to each.

**Q Gender.10**
Please provide a brief theoretical explanation in a narrative paragraph for your choices. Explain the assumptions or reasoning behind your beliefs about why these interactions are most important and why they take the signs you indicated.

Here, the survey displayed a text input box

## ⚠️ Important Notice Before You Continue

After you proceed to the next pages, you will no longer be able to return to this phase or modify your choices on the feature importance and signs.

Are you ready to continue?

**Predicting Mentions of Racial Inequality - Phase 2: Integrating Machine Learning Evidence**

In this phase, you will review the machine-learning results on **racial** inequality. You are then invited to refine or expand your theory in light of this new information.

**Machine Learning Evidence (1) - Main Effects**

Using the 13 features we've shown before, we trained a series of machine-learning models to predict whether a paper discusses **racial** inequality or not.

Below we report the top five predictive features for whether a paper discusses **racial** inequality, identified by Random Forest via Gini index, together with the sign of association estimated by Logistic Regression.

**What is the Gini index?** The Gini index is an impurity measure used by tree-based models such as Random Forests to assess how well a feature splits the data. A lower Gini index indicates that a split creates more homogeneous groups in terms of belonging or not to the positive class, meaning the feature is more useful for class prediction.

| Rank | Feature | Sign | Interpretation |
|---|---|---|---|
| 1 | social_science | + | Papers in social science fields are **more likely** to discuss racial inequality. |
| 2 | female_score | + | Papers with a higher share of female authors are **more likely** to discuss racial inequality. |
| 3 | country_race_diversity_score | + | Papers whose authors are inferred to be born in more racially diverse countries are **more likely** to discuss racial inequality. |
| 4 | asian | + | Papers with a higher estimated share of Asian authors are **more likely** to discuss racial inequality. |
| 5 | black | + | Papers with a higher estimated share of Black authors are **more likely** to discuss racial inequality. |

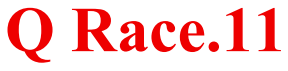

**Q Race.11**

**Please describe your reaction after viewing the ML results.** If the ML results diverge from your expectations, please explain why you think that might be the case.

Here, the survey displayed a text input box

**Q Race.12**

**Revise your theoretical explanation.**

Reminder of your initial theory for predicting whether a paper discusses **racial** inequality before ML evidence.

Here, the survey displayed the theoretical explanation provided by respondent in Q Race.4.

In light of the ML empirical evidence, you now have the opportunity to refine or update your previous theory on the **main effects** for predicting whether a paper discusses **racial** inequality.

Please provide a **complete** version of your updated theory, rather than only the modified parts, so that we can analyze your reasoning in a structured way. Your explanation should be a narrative paragraph in English.

**[Important Note]** If, after reviewing the ML results, you still believe your initial theory holds exactly as stated, you may choose not to make any modifications. However, if you decide not to update your theory, please explicitly indicate this in the text box and briefly explain why.

Here, the survey displayed a text input box

**Q Race.13**

**Please provide an updated diagram to represent your updated theory.**

Here, the survey displayed the diagram provided by respondent in Q Race.5.

**[Important Note]** If, after reviewing the ML results, you still believe your initial diagram holds exactly as presented, you may choose not to make any modifications. However, if you decide not to update your diagram, please explicitly indicate this in the text box.

Here, the survey displayed a text input box

**Machine Learning Evidence (2) - Second-Order Interactions**

Below we report the top three predictive second-order interactions for whether a paper discusses **racial** inequality, identified by Random Forest via Gini index, and the sign estimated by Logistic Regression.

| Rank | Second-Order Interaction | Sign | Interpretation |
|---|---|---|---|
| 1 | **social_science * paper_inequality_mentions_3_years** | + | • The marginal effect of recent academic attention to inequality on whether a paper discusses **racial** inequality is **larger** for social science papers.<br>• The marginal effect of being a social science paper on whether a paper discusses **racial** inequality is **larger** when recent academic attention to inequality is higher. |
| 2 | **social_science * news_inequality_mentions_3_years** | + | • The marginal effect of recent news attention to inequality on whether a paper discusses **racial** inequality is **larger** for social science papers.<br>• The marginal effect of being a social science paper on whether a paper discusses **racial** inequality is **larger** when recent news attention to inequality is higher. |
| 3 | **female * social_science** | - | • The marginal effect of female authorship on whether a paper discusses **racial** inequality becomes **smaller** for social science papers.<br>• The marginal effect of being a social science paper on whether a paper discusses **racial** inequality is **smaller** when the share of female authors is higher. |

**Q Race.14**

**Please describe your reaction after viewing the ML results.** If the results diverge from your expectations, please share why you think that might be the case.

Here, the survey displayed a text input box

**Q Race.15**

**Revise your theoretical explanation.**

Reminder of your initial theory on **second-order interactions** for predicting whether a paper discusses **racial** inequality.

Here, the survey displayed the theoretical explanation provided by respondent in Q Race.10.

In light of the ML empirical evidence, you now have the opportunity to refine or update your previous theory on the **second-order interactions** for predicting whether a paper discusses **racial** inequality.

Please provide a **complete** version of your updated theory (rather than only the revised parts) in a narrative paragraph.

**[Important Note]** If, after reviewing the ML results, you still believe your initial theory holds exactly as stated, you may choose not to make any modifications. However, if you decide not to update your theory, please explicitly indicate this in the text box and briefly explain why.

Here, the survey displayed a text input box

**Predicting Mentions of Racial Inequality - Phase 3: Empirical Tests**

In this phase, you will be asked to propose new empirical tests. These tests should focus on the main effects of predictors of whether a paper discusses **racial** inequality.

Reminder of your initial theoretical reasoning on main effects before ML evidence.

Here, the survey displayed the initial theoretical explanation and diagram provided by respondent in Q Race.4 and Q Race.5.

Reminder of your revised theoretical reasoning on main effects after ML evidence.

Here, the survey displayed the revised theoretical explanation and diagram provided by respondent in Q Race.12 and Q Race.13.

**Q Race.16**

**How could your revised theoretical reasoning on racial inequality be empirically tested using the available data?** Your proposed empirical tests may take the form of:

- a regression-based correlation analysis,
- a natural experiment,
- or other method you consider appropriate.

Please also describe what kinds of results would support your theoretical explanation.

**Important:** The empirical tests you propose must only use the 13 features available. In addition, the dataset covers publications from the years 2000 to 2020, and includes year_of_publication as a meta-feature for each paper. You are free to use this variable to code time-based events when proposing tests based on natural experiments (e.g., defining an indicator for whether a paper was published before vs. after a major societal event).

Here, the survey displayed a text input box

**Q Race.17**

**How could your theoretical reasoning be further tested using additional variables that you consider relevant but are NOT currently included in our list?** Please describe:

- additional variables that would strengthen the evaluation of your theory and how they could be used in new empirical tests, and
- expected results that would support your theory.

Here, the survey displayed a text input box

**Predicting Mentions of Gender Inequality - Phase 2: Integrating Machine Learning Evidence**

In this phase, you will review the machine-learning results on gender inequality. You are then invited to refine or expand your theory in light of this new information.

**Machine Learning Evidence (1) - Main Effects**

Using the 13 features we've shown before, we trained a series of machine-learning models to predict whether a paper discusses gender inequality or not.

Below we report the top five predictive features for whether a paper discusses gender inequality, identified by Random Forest via Gini index, together with the sign of association estimated by Logistic Regression.

**What is the Gini index?** The Gini index is an impurity measure used by tree-based models such as Random Forests to assess how well a feature splits the data. A lower Gini index indicates that a split creates more homogeneous groups in terms of belonging or not to the positive class, meaning the feature is more useful for class prediction.

| Rank | Feature | Sign | Interpretation |
|---|---|---|---|
| 1 | social_science | + | Papers published in social science fields are **more likely** to discuss gender inequality. |
| 2 | female_score | + | Papers with a higher share of female authors are **more likely** to discuss gender inequality. |
| 3 | natural_science | - | Papers published in natural science fields are **less likely** to discuss gender inequality. |
| 4 | asian | + | Papers with a higher estimated share of Asian authors are **more likely** to discuss gender inequality. |
| 5 | paper_inequality_mentions_3_years | + | Papers published during periods when a higher share of academic work addresses inequality are **more likely** to discuss gender inequality. |

**Q Gender.11**

**Please describe your reaction after viewing the ML results.** If the ML results diverge from your expectations, please explain why you think that might be the case.

Here, the survey displayed a text input box

**Q Gender.12**

**Revise your theoretical explanation.**

Reminder of your initial theory for predicting whether a paper discusses **gender** inequality before ML evidence.

Here, the survey displayed the theoretical explanation provided by respondent in Q Gender.4.

In light of the ML empirical evidence, you now have the opportunity to refine or update your previous theory on the **main effects** for predicting whether a paper discusses **gender** inequality.

Please provide a **complete** version of your updated theory, rather than only the modified parts, so that we can analyze your reasoning in a structured way. Your explanation should be a narrative paragraph in English.

**[Important Note]** If, after reviewing the ML results, you still believe your initial theory holds exactly as stated, you may choose not to make any modifications. However, if you decide not to update your theory, please explicitly indicate this in the text box and briefly explain why.

Here, the survey displayed a text input box

**Q Gender.13**

**Please provide an updated diagram to represent your updated theory.**

Here, the survey displayed the diagram provided by respondent in Q Gender.5.

**[Important Note]** If, after reviewing the ML results, you still believe your initial diagram holds exactly as presented, you may choose not to make any modifications. However, if you decide not to update your diagram, please explicitly indicate this in the text box.

Here, the survey displayed a text input box

**Machine Learning Evidence (2) - Second-Order Interactions**

Below we report the top three predictive second-order interactions for whether a paper discusses gender inequality, identified by Random Forest via Gini index, and the sign estimated by Logistic Regression.

| Rank | Second-Order Interaction | Sign | Interpretation |
|---|---|---|---|
| 1 | social_science * paper_inequality_mentions_3_years | + | • The marginal effect of recent academic attention to inequality on whether a paper discusses gender inequality is **larger** for social science papers.<br>• The marginal effect of being a social science paper on whether a paper discusses gender inequality is **larger** when recent academic attention to inequality is higher. |
| 2 | social_science * news_inequality_mentions_3_years | + | • The marginal effect of recent news attention to inequality on whether a paper discusses gender inequality is **larger** for social science papers.<br>• The marginal effect of being a social science paper on whether a paper discusses gender inequality is **larger** when recent news attention to inequality is higher. |
| 3 | female * social_science | - | • The marginal effect of female authorship on whether a paper discusses gender inequality becomes **smaller** for social science papers.<br>• The marginal effect of being a social science paper on whether a paper discusses gender inequality is **smaller** when the share of female authors is higher. |

**Q Gender.14**

**Please describe your reaction after viewing the ML results.** If the results diverge from your expectations, please share why you think that might be the case.

Here, the survey displayed a text input box

**Q Gender.15**

**Revise your theoretical explanation.**

Reminder of your initial theory on **second-order interactions** for predicting whether a paper discusses **gender** inequality.

Here, the survey displayed the theoretical explanation provided by respondent in Q Gender.10.

In light of the ML empirical evidence, you now have the opportunity to refine or update your previous theory on the **second-order interactions** for predicting whether a paper discusses **gender** inequality.

Please provide a **complete** version of your updated theory (rather than only the revised parts) in a narrative paragraph.

**[Important Note]** If, after reviewing the ML results, you still believe your initial theory holds exactly as stated, you may choose not to make any modifications. However, if you decide not to update your theory, please explicitly indicate this in the text box and briefly explain why.

Here, the survey displayed a text input box

**Predicting Mentions of Gender Inequality - Phase 3: Empirical Tests**

In this phase, you will be asked to propose new empirical tests. These tests should focus on the main effects of predictors of whether a paper discusses **gender** inequality.

Reminder of your initial theoretical reasoning on main effects before ML evidence.

Here, the survey displayed the initial theoretical explanation and diagram provided by respondent in Q Gender.4 and Q Gender.5.

Reminder of your revised theoretical reasoning on main effects after ML evidence.

Here, the survey displayed the revised theoretical explanation and diagram provided by respondent in Q Gender.12 and Q Gender.13.

**Q Gender.16**

**How could your revised theoretical reasoning on gender inequality be empirically tested using the available data?** Your proposed empirical tests may take the form of:

- a regression-based correlation analysis,
- a natural experiment,
- or other method you consider appropriate.

Please also describe what kinds of results would support your theoretical explanation.

**Important:** The empirical tests you propose must only use the 13 features available. In addition, the dataset covers publications from the years 2000 to 2020, and includes year_of_publication as a meta-feature for each paper. You are free to use this variable to code time-based events when proposing tests based on natural experiments (e.g., defining an indicator for whether a paper was published before vs. after a major societal event).

Here, the survey displayed a text input box

### Q Gender.17

**How could your theoretical reasoning be further tested using additional variables that you consider relevant but are NOT currently included in our list?** Please describe:

- additional variables that would strengthen the evaluation of your theory and how they could be used in new empirical tests, and
- expected results that would support your theory.

Here, the survey displayed a text input box

**Final Step**

As part of this study, GenAI models will perform the exact same full sequence of theory-generation tasks that you completed. We are interested in your expectations regarding the performance of different sources of theories.

Which of the following sets of proposed theories do you think will be rated highest in quality (e.g., clarity, creativity, coherence, originality, rigor, and overall persuasiveness) by independent researchers? "Independent researchers" here refers to a *separate group of human evaluators* who are not part of this survey and do not know whether the theories were generated by humans or by GenAI.

- Theories proposed by human experts after reviewing the machine-learning results
- Theories proposed by GenAl models after reviewing the machine-learning results
- Theories proposed by GenAI models after reviewing (1) the machine-learning results and (2) the revised theories of human experts who have reviewed the machine-learning results

Which of the following sets of proposed theories do you think will receive the strongest empirical support when tested (assuming each theory is tested with the empirical tests that were proposed to test that theory)?

- Theories proposed by human experts after reviewing the machine-learning results
- Theories proposed by GenAl models after reviewing the machine-learning results
- Theories proposed by GenAI models after reviewing (1) the machine-learning results and (2) the revised theories of human experts who have reviewed the machine-learning results

As a key contributor and future co-author in shaping this human–AI collaborative framework, your insights are truly invaluable. Please provide any additional comments, critiques, or thoughts you would like to share with us.

Here, the survey displayed a text input box

**Demographic Information**
The following questions ask for demographic and professional background information.

- What is your age?
- What is your gender?
- In which country/region were you born?
- In which country/region do you currently reside?
- How many years of experience with English do you have?
- What year did you receive your doctoral degree, or in what year do you expect to receive it?
- What is your department or academic field at your institution (e.g., psychology, organizational behavior, computer science, statistics)?
- What is your job rank?
- Approximately how many peer-reviewed academic articles have you published on topics related to racial inequality (e.g., racial discrimination, race gaps, racial stereotyping, or racial bias)?
- Approximately how many peer-reviewed academic articles have you published on topics related to gender inequality (e.g., gender discrimination, gender gaps, gender stereotyping, or gender bias)?
- Approximately how many times have you taught (including being a TA for) a graduate-level statistics or methods course (e.g., research design, quantitative/qualitative methods)?

**Co-authorship**
Completing the entire survey qualifies you to be listed as a co-author on the manuscript reporting the results. Would you like to be listed as a co-author?

If yes,

- first name as you would like it to appear on the manuscript:
- last name as you would like it to appear on the manuscript:
- (optional) Middle initial as you would like it to appear on the manuscript:
- institutional affiliation as you would like it to appear on the manuscript: